\documentclass[10pt,journal,compsoc]{IEEEtran}
\usepackage{graphicx}
\usepackage{amsmath,amssymb} 
\usepackage{color}
\usepackage{todonotes}

\usepackage{algorithm}
\usepackage{algpseudocode}  
\usepackage{multirow}
\usepackage{booktabs}
\usepackage{subfigure}
\usepackage{hyperref}
\usepackage{amsmath}
\usepackage{cases}
\usepackage{makecell}
\usepackage{pifont}
\usepackage{xcolor}

\ifCLASSOPTIONcompsoc
  \usepackage[nocompress]{cite}
\else
  \usepackage{cite}
\fi
\ifCLASSINFOpdf
\else
\fi
\begin{document}
%

\title{GRATIS: Deep Learning \textbf{G}raph \textbf{R}epresentation with T\textbf{a}sk-specific \textbf{T}opology and Mult\textbf{i}-dimensional Edge Feature\textbf{s}}
%
%
%
%

\author{Siyang Song,
        Yuxin Song,        
        Cheng Luo,
        Zhiyuan Song,
        Selim Kuzucu,
        Xi Jia,
        Zhijiang Guo,
        Weicheng Xie,
        Linlin Shen,
        and~Hatice Gunes
        
\IEEEcompsocitemizethanks{\IEEEcompsocthanksitem Siyang Song, Zhiyuan Song, Zhijiang Guo and Hatice Gunes are with the Department of Computer Science and Technology, University of Cambridge, Cambridge, CB3 0FT, United Kingdom.
E-mail: ss2796@cam.ac.uk, zs393@cam.ac.uk, zg283@cam.ac.uk, hatice.gunes@cl.cam.ac.uk

\IEEEcompsocthanksitem Yuxin Song is with the Department of Computer Vision Technology (VIS), Baidu Inc, Beijing 100193, China. (E-mail: songyuxin02@baidu.com)

\IEEEcompsocthanksitem Cheng Luo, Weicheng Xie and Linlin Shen are with Computer Vision Institute, School of Computer Science \& Software Engineering, Shenzhen Institute of Artificial Intelligence and Robotics for Society, Guangdong Key Laboratory of Intelligent Information Processing, Shenzhen University, Shenzhen, 518060, China. (Corresponding author: Prof. Linlin Shen, llshen@szu.edu.cn)

\IEEEcompsocthanksitem Selim Kuzucu is with Department of Computer Engineering, Middle East Technical University,  Ankara, Turkey.

\IEEEcompsocthanksitem Xi Jia is with School of Computer Science, University of Birmingham,  Birmingham, UK.}

\thanks{Manuscript received September 19, 2022.}}

%
%

\markboth{Journal of \LaTeX\ Class Files,~Vol.~14, No.~8, August~2015}%
{Shell \MakeLowercase{\textit{et al.}}: Bare Advanced Demo of IEEEtran.cls for IEEE Computer Society Journals}
%



\IEEEtitleabstractindextext{%
\begin{abstract}

Graph is powerful for representing various types of real-world data. The topology (edges' presence) and edges' features of a graph decides the message passing mechanism among vertices within the graph. While most existing approaches only manually define a single-value edge to describe the connectivity or strength of association between a pair of vertices, task-specific and crucial relationship cues may be disregarded by such manually defined topology and single-value edge features. In this paper, we propose the first general graph representation learning framework (called GRATIS) which can generate a strong graph representation with a task-specific topology and task-specific multi-dimensional edge features from any arbitrary input (i.e., graph or non-graph data). To learn each edge's presence and multi-dimensional feature, our framework takes both of the corresponding vertices pair and their global contextual information into consideration, enabling the generated graph representation to have a globally optimal message passing mechanism for different down-stream tasks. The principled investigation results achieved for various graph analysis tasks (e.g., graph classification, vertex classification, and link prediction) on 11 graph and non-graph datasets show that our GRATIS can not only largely enhance pre-defined graphs but also learns a strong graph representation for non-graph data, with clear performance improvements on all tasks. In particular, the learned topology and multi-dimensional edge features provide complementary task-related cues for graph analysis tasks. Our framework is effective, robust and flexible, and is a plug-and-play module that can be combined with different backbones and Graph Neural Networks (GNNs) to generate a task-specific graph representation from various graph and non-graph data. Our code is made publicly available at \url{https://github.com/SSYSteve/Learning-Graph-Representation-with-Task-specific-Topology-and-Multi-dimensional-Edge-Features}.


\end{abstract}

}

\maketitle

\IEEEdisplaynontitleabstractindextext

%
\IEEEpeerreviewmaketitle

\ifCLASSOPTIONcompsoc
\IEEEraisesectionheading{\section{Introduction}\label{sec:introduction}}
\else
\section{Introduction}
\label{sec:introduction}
\fi


\begin{figure}
	\centering
	\subfigure[Comparison for representing graph data. (i) Many existing approaches \cite{wang2021graphtcn,weng2020gnn3dmot,xu2021two} only represent the relationship between each pair of vertices using a binary value, describing whether they are associated; (ii) some approaches \cite{shao2020spatio,isufi2021edgenets} using a single value-edge to describe the strength of association between each pair of vertices; (iii) \textbf{GRATIS} not only assigns a task-specific topology by adding task-specific edges (depicted in red) for the graph but also describes multiple task-specific relationship cues between each pair of connected vertices using a multi-dimensional edge feature.]{\label{subfig:comparsion_graph}
		\includegraphics[width=9cm]{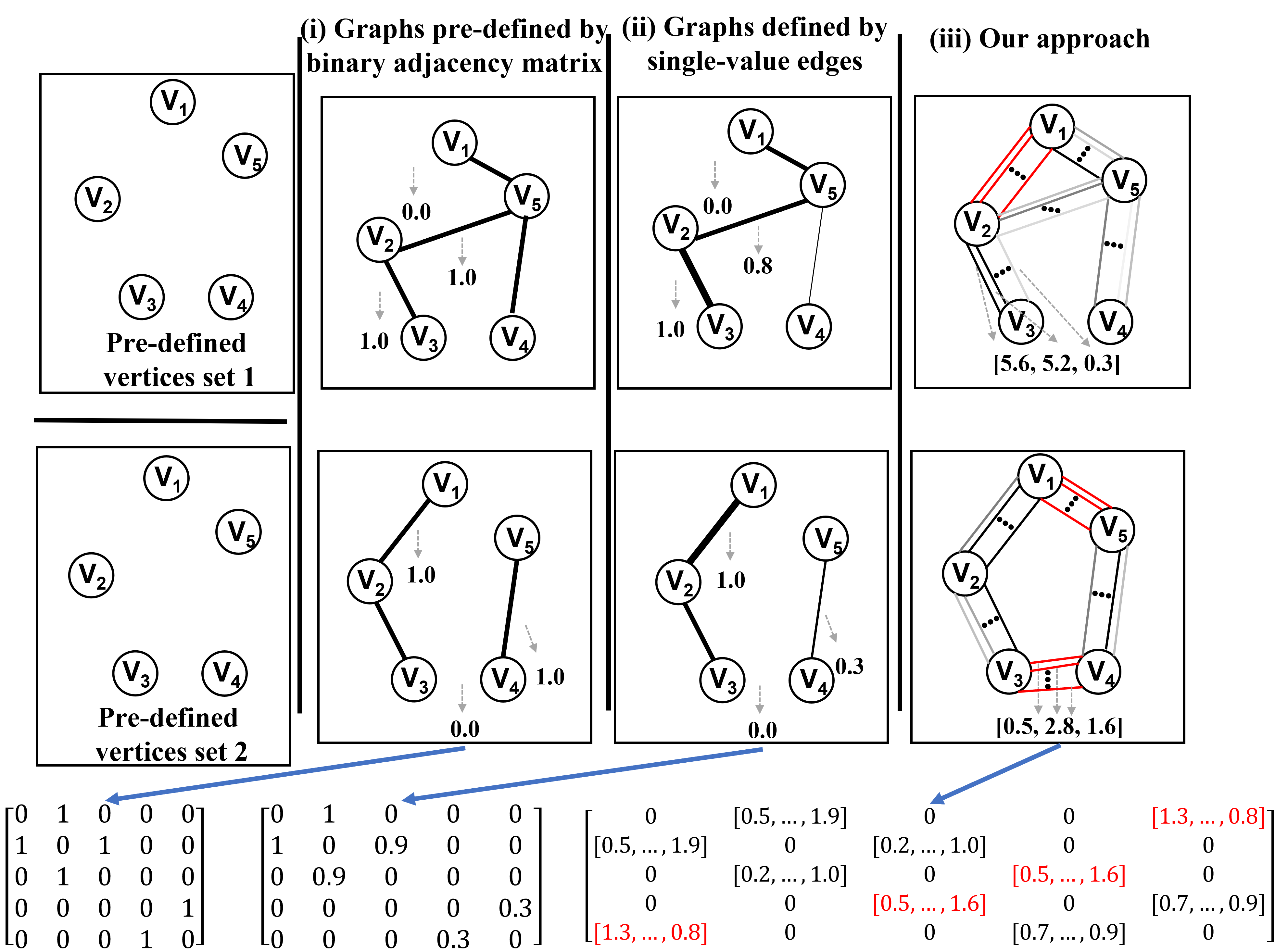}}
	\subfigure[Comparison for representing non-graph data, where the example is using graphs to representing facial Action Units (AUs). (i) \textbf{pre-defined AU graphs} that use a single topology to define AU association for all facial displays \cite{li2019semantic,liu2020relation}; (ii) \textbf{Facial display-specific AU graphs} that assign a unique topology to define AU association for each facial display \cite{song2021uncertain,Song_2021_CVPR}. Both (i) and (ii) use a single value as an edge feature; (iii) \textbf{GRATIS} encodes a unique AU association pattern for each facial display in node features which also decide the task-specific topology of the graph, and additionally describes each edge (the relationship between a pair of AUs) using a multi-dimensional feature.]{\label{subfig:comparsion-nongraph}
	\includegraphics[width=9cm]{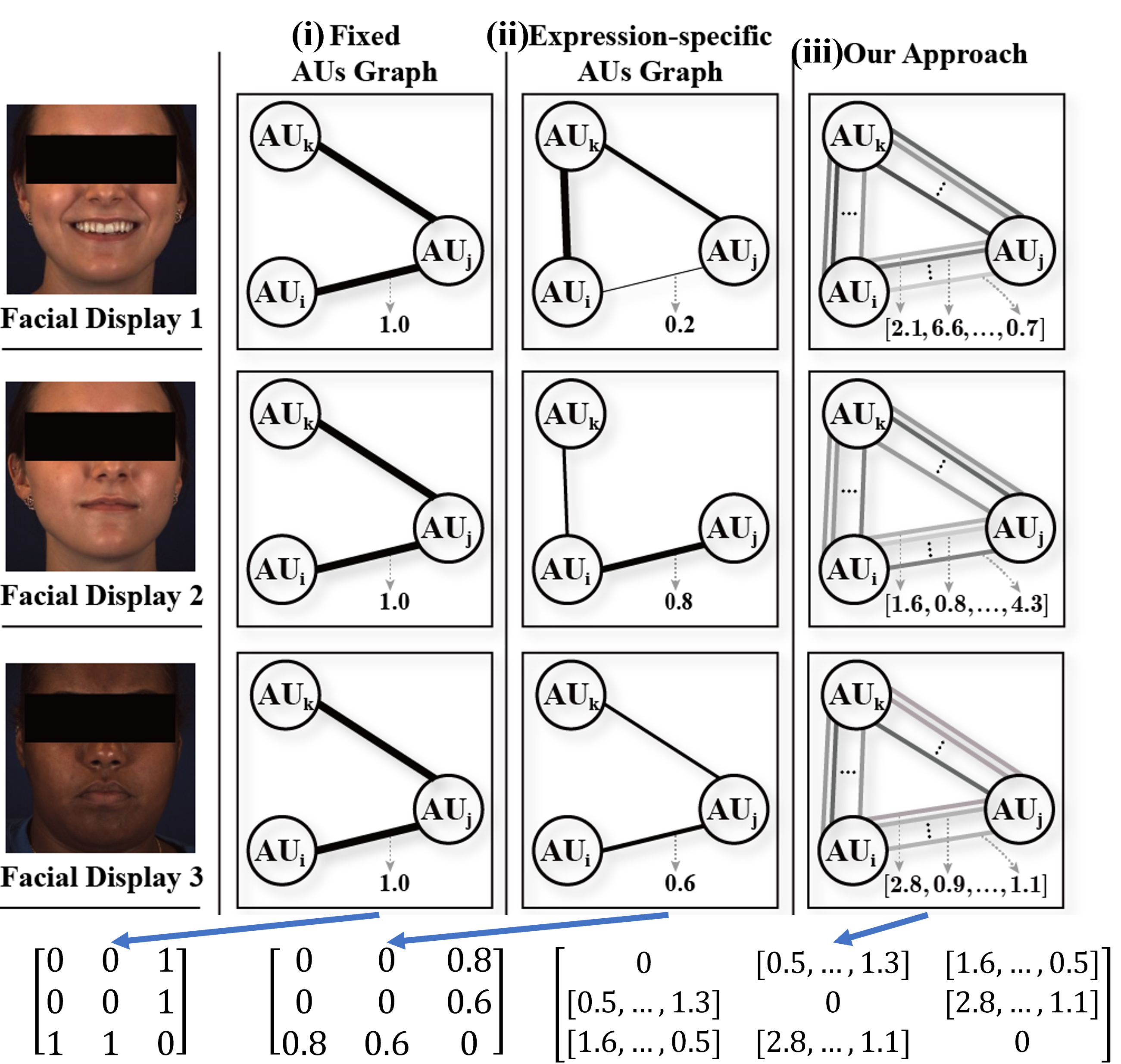}}
	\caption{Comparison between the graph representations generated by our GRATIS and existing approaches, where edge representations of graphs compared in the last row are also provided. Our approach is the first framework that can generate a graph representation with task-specific topology and multi-dimensional edge features from any graph or non-graph data.} 
    \label{fig:comparsion-graph-nongraph}
\end{figure}

\noindent \IEEEPARstart{G}{raphs} have been widely recognized as powerful representations to describe a wide variety of real-world data such as images \cite{song2021uncertain,li2019semantic}, social relationships \cite{yanardag2015deep,giles1998citeseer,sen2008collective}, and human skeleton \cite{yan2018spatial}. A typical graph consists of a set of vertices and edges, where each vertex usually represents the mathematical abstraction of an object, and each edge describes the relationship between a pair of vertices \cite{west2001introduction}. To encode a raw data sample as a graph, the majority of existing approaches \cite{dwivedi2020benchmarking,abbe2017community,giles1998citeseer,yan2018spatial} manually define the vertices, the topology of the graph (i.e., edges' presence) as well as each edge's features based on pre-defined rules. Specifically, most of them use either a binary value \cite{yan2018spatial,wang2021graphtcn,lo2020mer,weng2020gnn3dmot,xu2021two} or a single-value weight \cite{shao2020spatio,isufi2021edgenets} as each edge's representation to describe the connectivity or strength of association between a pair of vertices.

While the graph topology defined by a pre-defined rule can only represent a specific relationship pattern among vertices, some related vertex pairs whose relationships are not considered by the rule would be treated as 'un-related', i.e., these vertices would not be connected in the manually-defined graph. Such hand-crafted strategies \cite{li2019semantic,zhang2020region,liu2020relation,xu2021two} frequently assign the same topology for graph representations of all samples in the dataset. Subsequently, task-related connections may be ignored in the manually-defined graphs, and thus the performance of the graph analysis would be limited. Although some recent studies propose to learn the association strength or connectivity between vertices using the target for supervision \cite{weng2020gnn3dmot,wang2021graphtcn,ioannidis2019recurrent,song2021uncertain,lei2020novel}, these methods were only proposed to generate task-specific graph typologies for graph representations of a specific type of non-graph data (e.g., face image \cite{song2021uncertain}, human skeleton \cite{weng2020gnn3dmot}, etc.). In other words, none of them can be applied to multiple types of non-graph data nor pre-defined graphs.

Edge features are also essential components of graphs \cite{hamilton2020graph}. While the relationship between a pair of connected vertices sometimes can be described by multiple attributes if they are not linearly dependent (e.g., the relationship between a pair of people can be described by the differences between their ages, gender and nationalities, etc.), the majority of existing graphs \cite{giles1998citeseer,mccallum2000automating} only employ a single value as each edge's feature to describe their relationships, which usually ignore crucial relationship cues. To comprehensively utilize rich relationship cues between vertices for graph analysis tasks, several studies \cite{gong2019exploiting,jo2021edge,vashishth2019composition} developed novel edge message passing methods that allow GNNs to process multi-dimensional edge feature-based graphs. However, instead of creating a multi-dimensional feature for representing each edge, these methods only focus on efficiently processing multi-dimensional edge features that are already contained in the input graph. Although some studies manually design multi-dimensional edge features to describe some specific relationship between vertices, e.g., distances \cite{li2020customized}, latent interactions \cite{mavroudi2020representation}, heuristic information \cite{wang2020learning}, etc., these hand-crafted edge features still fail to learn task-specific relationship cues between vertices. In summary, there is lack of a generic graph representation learning framework that can automatically generate a graph representation that has a task-specific topology and multi-dimensional edge features, for any arbitrary input data (i.e., pre-defined graph or non-graph data such as image, video, audio, and text).



\begin{figure}
	\centering
	\subfigure[The pipeline for processing pre-defined graphs]{\label{subfig:graph_data}
		\includegraphics[width=9cm]{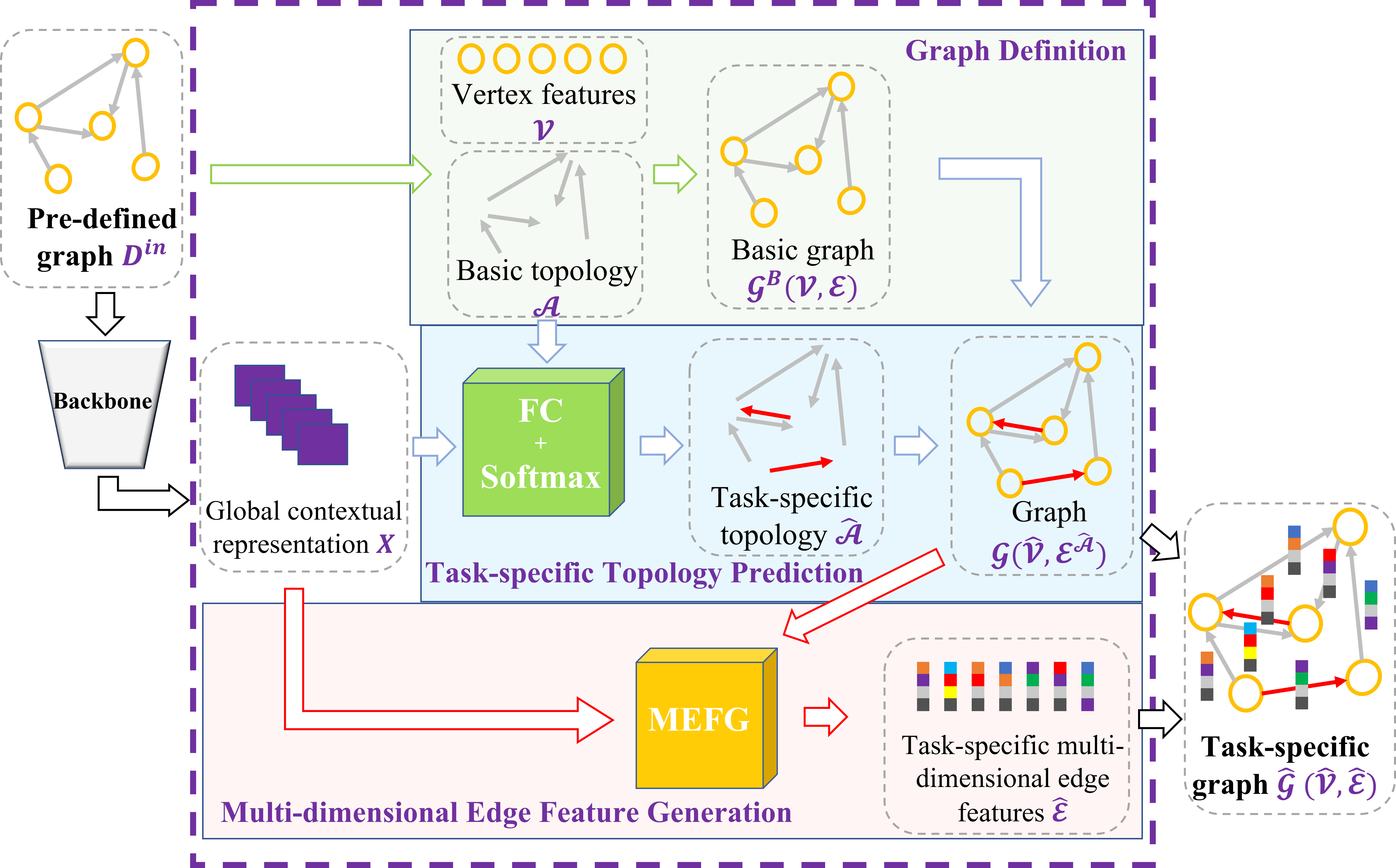}}
	\subfigure[The pipeline for processing non-graph data]{\label{subfig:non-graph_data}
		\includegraphics[width=9cm]{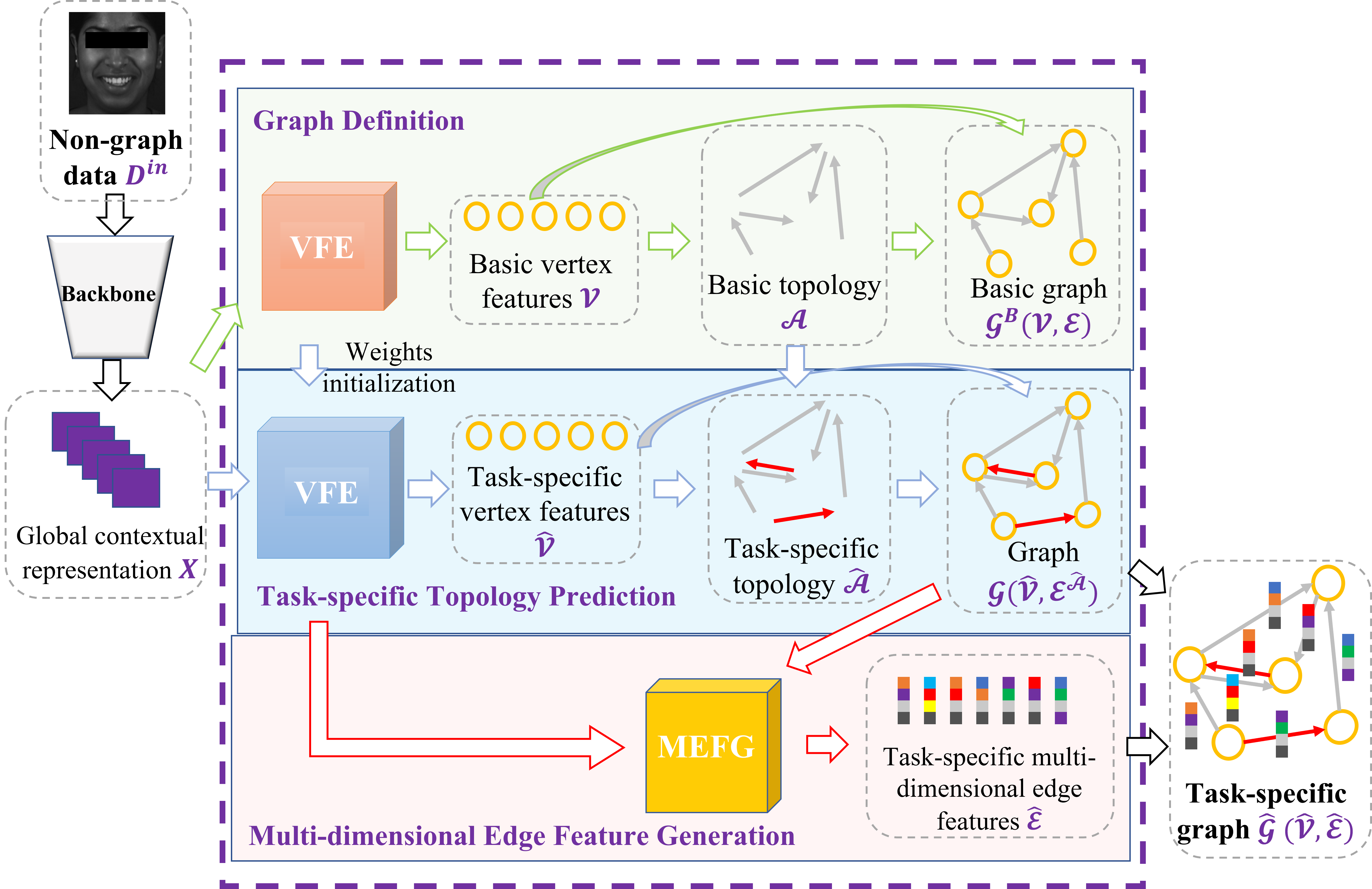}}
	\caption{The pipeline of the proposed GRATIS. The displayed two sub-figures attempt to produce a graph representation for a pre-defined graph (sub-figure (a)) or a non-graph data (sub-figure (b)), respectively. Both pipelines consist of four main steps: (i) global contextual representation $X$ is generated from the backbone; (ii) Graph Definition (depicted in green, and explained in Sec. \ref{subsec:approach-graph definition}) that defines a basic graph representation for the input data; (iii) Task-specific Topology Prediction (depicted in blue, and explained in Sec. \ref{subsec:approach-topology}); and (iv) Multi-dimensional Edge Feature Generation (depicted in red, and explained in Sec. \ref{subsec:approach-multi-dimensional edge feature}). Here, the VFE denotes the Vertex Feature Extraction network employed in the non-graph data-based framework.} 
    \label{fig:method_overview}
\end{figure}

To bridge the research gaps described above, in this paper, we propose \textbf{the first generic graph representation learning framework} which can \textbf{generate a task-specific graph representation with a task-specific topology and multi-dimensional edge features from an arbitrary graph or non-graph data}. The proposed framework can be easily combined with various deep learning backbones and GNN predictors for different down-stream tasks. The proposed GRATIS consists of three modules: a \textbf{Graph Definition (GD) module}, a \textbf{Task-specific Topology Prediction (TTP) module} and a \textbf{Multi-dimensional Edge Feature Generation (MEFG) module}. Specifically, the GD module first defines a basic graph representation (i.e., vertex features, basic topology, and edge features) from the output of the backbone, which learns a set of task-specific vertex features for non-graph data. Then, the TTP module generates a task-specific adjacency matrix that allows the produced graph to better represent relationship cues carried by the input data (it also generates a set of task-specific vertex features for the graph representation of non-graph data). Finally, the MEFG module assigns a multi-dimensional feature for each presented edge, describing multiple relationship cues between the corresponding pair of vertices. Since we train the GD, TTP, and MEFG with the backbone and GNN predictor in an end-to-end manner, both TTP and MEFG are learned to assign task-specific topology and edge features for the final produced graph representation. Moreover, the TTP and MEFG considers not only the relationship between corresponding vertex features but also the global contextual information of both vertices, to decide each edge's presence and feature. As a result, the generated topology and multi-dimensional edge features are expected to lead the graph to have a globally optimal message passing mechanism within the graph for the down-stream task. The comparison between the our approach to other related studies are illustrated in Fig. \ref{fig:comparsion-graph-nongraph}. The details of the proposed framework are visualized in Fig. \ref{fig:method_overview}. The main contributions of this paper are summarized as follows:
\begin{itemize}




    \item We propose the first task-specific graph topology learning framework that can not only enhance any graph with a pre-defined topology to a graph that has a task-specific topology, but also automatically generate a graph representation that containing task-specific vertex features and topology to represent a non-graph data, where each vertex feature is learned to not only represent the corresponding object but also its association with other objects (vertices).

    \item We propose the first general strategy that can assign task-specific multi-dimensional edge features for any arbitrary graph with either single-value or multi-dimensional edge features. Importantly, it produces a multi-dimensional feature for each presented edge by considering not only the corresponding vertices pair but also the global context, to describe multiple task-specific relationship cues between the corresponding pair of vertices.

    \item We evaluate the proposed plug-and-play framework on three standard graph analysis tasks (e.g., graph classification, node classification, and link prediction) using 10 graph and non-graph (facial image) datasets. The results demonstrate that the task-specific multi-dimensional edge graph representations generated by our framework consistently and robustly enhanced the analysis performance for various tasks under different settings (e.g., backbone models, models settings, types of data, and GNN predictors).

\end{itemize}
In comparison to the previous conference version \cite{luo2022learning}, the additional contributions of this paper are: \textbf{Methodology:} (i) While the previous version only learns a graph representation from a face image for AU recognition (i.e., non-graph vertex classification), this paper further extends it as a general plug-and-play framework that can learn graph representations from various graph and non-graph data for different graph analysis (i.e., graph classification, vertex classification, and link prediction); and (ii) Based on the previous strategy, this paper proposes new topology learning and multi-dimensional edge feature learning strategies, allowing the proposed framework to learn task-specific graph representations from pre-defined graphs. \textbf{Experiment:} This paper additionally: (i) employed the GAT as the GNN predictor, and evaluated all systems based on it; (ii) conducted non-graph data-based graph classification (the FER and depression recognition tasks) and link prediction (the AU co-occurrence prediction task) experiments, based on two different backbones and two different GNN predictors; (iii) conducted graph-based graph classification, vertex classification and link prediction experiments based on two GNN predictors; (iv) conducted a set of ablation studies to evaluate the proposed TTP and MEFG modules under different tasks; and (v) evaluated multiple ways of utilising vertex and edge features for link prediction tasks.

\section{Related Work}
\label{sec:related work}

\noindent In this section, we focus on reviewing edges' presence and feature generation strategies that have been used by previous graph-based approaches and applications, which are categorized as single-value edge feature-based approaches and multi-dimensional edge feature-based approaches.

\subsection{Graph topology generation and single-value edge feature}

\noindent The majority of existing studies use a binary adjacency matrix to define the topology of a graph, i.e., the connectivity/edge presence between vertices. Some widely-used graph analysis datasets (Cora \cite{mccallum2000automating}, PubMed \cite{sen2008collective} and CiteSeer \cite{giles1998citeseer}) and link prediction datasets (PATTERN and CLUSTER \cite{dwivedi2020benchmarking}) are typical examples. In these datasets, each publication is represented as a vertex. Meanwhile, the citation relationship between publications, category of vertices, or collaboration relationship between vertices \cite{yanardag2015deep} have been employed to decide the edge presence for each pair of vertices. Besides pre-defined graphs, recent studies frequently represent various types of non-graph data as graphs, where binary-value edges are also employed. Yan et al. \cite{yan2018spatial} manually define a binary adjacency matrix that connects adjacent human body skeleton points (spatial edges) within each frame as well as the same skeleton points in adjacent frames (temporal edges). A similar strategy also has been frequently employed to represent facial landmarks sequences \cite{chen2019efficient,abbasi2022statistical}. In \cite{liu2020relation}, features extracted from each Facial Action Unit (AU)-related region \cite{friesen1978facial} are represented as a vertex, and each edge's feature is determined by the conditional co-occurrence probability of a pair of AUs, which is a binary value. In addition, some studies also compute a real value such as distance (Hop-distance or Euclidean distance) \cite{zhou2020facial} or Pearson Correlation Coefficient (PCC) \cite{shao2020spatio} between vertex features to represent each edge, rather than using a binary value. 


Instead of using a pre-defined adjacency matrix, Liu et al. \cite{liu2021video} propose to learn a graph to represent facial image sequence, where the adjacency matrix of the graph is deep learned based on the relationship among facial regions in the sequence. Weng et al. \cite{weng2020gnn3dmot} construct a spatio-temporal graph for an image sequence to track multiple 3D objects, which takes each detected objects as vertices while the edge presence between each pair of inter- or intra-frame vertices are decided by the distance between the corresponding deep-learned vertex features and the pre-defined distance thresholds in 2D and 3D spaces. Meanwhile, some studies attempt to learn a weight adjacency matrix to describe the strength of association between vertices, i.e., each edge is represented by a real value. Ioannidis et al. \cite{ioannidis2019recurrent} consider the relationship between each pair of vertices as a dynamic and non-linear process, which is adaptively obtained through a set of learnable weights. Isufi et al. \cite{isufi2021edgenets} propose an Edge Varying Graph Neural Network that deep learns a single-value weight for each edge, allowing each vertex to use a single value edge to influence each neighbouring vertex. Song et al. \cite{song2021uncertain} employ a similar mechanism as Graph Attention Network (GAT) \cite{velivckovic2017graph}, which deep learns a weighted mask to describe the importance of each edge (the strength of association between AUs) in a facial graph. Wang et al. \cite{wang2021graphtcn} use a graph to model pedestrians' interactions, where each edge is deep-learned to describe the relative spatial relationship between a pair of pedestrians. Zhang et al. \cite{zhang2019context} construct a facial affective graph for emotion recognition, where each local facial region is denoted as a vertex. This work deep learns a weight for each edge to measure the emotional intensity between a pair of local facial regions.

\subsection{Multi-dimensional edge feature}

\noindent Since a single-value edge usually can not explicitly describe the complex relationship between a pair of vertices, recent studies started to investigate the usage of multi-dimensional vector as an edge feature. As a result, novel message passing mechanisms that allow GNNs to process multi-dimensional edge features during the graph propagation have been widely investigated \cite{gong2019exploiting,jo2021edge,vashishth2019composition,kim2019edge,hussain2021edge,cai2021edge}. Meanwhile, the majority of existing approaches only manually define multi-dimensional edge features for graphs. Li et al. \cite{li2020customized} build a graph representation for analog IC placement, where devices are treated as vertices and their connections are represented as edges. This method computes several statistics of devices' attributes and the spatial distance between devices as the multi-dimensional edge feature. Mavroudi et al.\cite{mavroudi2020representation} manually define edge features that describes multiple aspects of interactions between actors and objects in the video. In \cite{mandya2020contextualised}, two types of multi-dimensional edge features are defined for graph-based relation extraction, which are dependency relation-based edge features and connection type edge features. Wang et al. \cite{wang2020learning} manually design multi-dimensional edge features for representing the underlying relationships between videos and texts in a video-text graph, which are initialized with heuristic information.

Since these hand-crafted multi-dimensional edge features are decided by pre-defined rules, they may fail to contain crucial task-specific relationship cues between vertices. Consequently, a small number of recent approaches started to learn task-specific multi-dimensional edge features for graphs. Bai et al. \cite{bai2020boundary} describe each video as a graph, which treats video segment boundaries (time stamps of the video) as vertices and the contents of video segments as edges. Consequently, a CNN is learned to produce a multi-dimensional feature from the video segment to describe each edge. Given a graph, Xiong et al. \cite{xiong2021multi} propose to deep learn a set of feature maps from each vertex feature, and each multi-dimensional edge feature is computed based on the similarity between the feature maps produced from the corresponding vertices. Aiming to identify local muscle movements, Lei et al \cite{lei2020novel} feed features of a pair of connected vertices (each vertex represents a local muscle movement) into a pre-trained CNN to produce an edge's feature, i.e., the relationship between a pair of local muscle movements. In \cite{shao2021personality,song2021learning}, the parameters and architecture of a person-specific network are encoded into a graph representation, where network layers are represented as vertices and each multi-dimensional edge feature is deep learned from a pair of adjacent vertices to describe the relationship between them. The experimental results demonstrate that such deep-learned multi-dimensional edge features clearly enhanced the graph regression performance over the manually defined single-value edge features. However, all the aforementioned approaches can be only applied to a specific task without generality. In addition, they only consider the corresponding vertex pair when generating a multi-dimensional edge feature, which ignores the crucial global contextual information.

In summary, although the aforementioned reviewed approaches already provide some topology and single-value edge feature learning strategies, to the best of our knowledge, there is no previous study provide a task-specific multi-dimensional graph edge feature learning solution. In other words, current GNNs can only process graphs that already contain multi-dimensional edge features, rather than automatically learn multi-dimensional edge features for constructing a graph representation from a non-graph data or a single-value edge graph. 



\section{Preliminaries}
\label{sec:Preliminary}

\noindent In this section, we briefly introduce basic concepts of the graph representation as well as the general vertex and edge updating mechanism of GNNs. 

\subsection{Graph representation} 

\noindent A graph $\mathcal{G} = (\mathcal{V}, \mathcal{E})$ is made up of a set of vertices $\mathcal{V} \subseteq \{\mathbf{v}_i \in \mathbb{R}^{1 \times K} \}$, and edges $\mathcal{E} \subseteq \{ \mathbf{e}_{i,j} = \mathbf{e}(\mathbf{v}_i, \mathbf{v}_j) \mid \mathbf{v}_i, \mathbf{v}_j \in \mathcal{V},  i \neq j \}$, where $\mathbf{v}_i$ represents $K$ attributes of the $i_{th}$ object in the graph/original data and $\mathbf{e}_{i,j}$ represents the edge feature that defines the relationship between vertices $\mathbf{v}_i$ and $\mathbf{v}_j$. Each pair of vertices can only be connected by at most one undirected edge or two directed edges. A standard way to describe such edges is through the adjacency matrix $\mathcal{A} \in \mathbb{R}^{|\mathcal{V}| \times |\mathcal{V}|}$, where all vertices in a graph are ordered so that each vertex indexes a specific row and column. As a result, the presence of each edge $\mathbf{e}_{i,j}$ can be described by a binary value $\mathcal{A}_{i,j} = 1$ if $\mathbf{v}_i$ and $\mathbf{v}_j$ are connected or $\mathcal{A}_{i,j} = 0$ otherwise. Specifically, the adjacent matrix is always symmetric if all edges are undirected, but can be non-symmetric if one or more directed edges exist. Instead of using a binary value, some studies \cite{dwivedi2020benchmarking,song2021uncertain,isufi2021edgenets} also build adjacency matrices with continuous real values to describe the strength of association between each pair of vertices.


\subsection{Vertex and edge updating mechanism of GNNs}

\noindent Recently, Graph Neural Networks (GNNs) (including Graph Convolution Networks (GCNs)) \cite{kipf2016semi,xu2018powerful,bresson2017residual,velivckovic2017graph} are dominant models that have been applied to a wide variety of graph data-based tasks. For a GNN $G$, its $l_{th}$ layer $G^l$ takes the graph $\mathcal{G}^{l-1} = (\mathcal{V}^{l-1}, \mathcal{E}^{l-1})$ that is produced by the ${l-1}_{th}$ layer as the input, and generates a new graph $\mathcal{G}^{l} = (\mathcal{V}^l, \mathcal{E}^l)$, which can be formulated as:
\begin{equation}
    \mathcal{G}^{l} = G^l(\mathcal{G}^{l-1})
\end{equation}
Specifically, the vertex feature $\mathbf{v}^l_i$ in $\mathcal{G}^{l}$ is computed based on: (i) its previous status $\mathbf{v}_i^{l-1}$ in $\mathcal{G}^{l-1}$; (ii) the adjacent vertices $\mathbf{v}_j^{l-1}$ of the $\mathbf{v}_i^{l-1}$ in $\mathcal{G}^{l-1}$ (denoted as the $\mathbf{v}_j^{l-1} \subseteq \mathcal{N}(\mathbf{v}_i^{l-1})$, where $\mathcal{A}^{l-1}_{i,j} = 1$ and $\mathcal{A}^{l-1}$ is the adjacent matrix of the $\mathcal{G}^{l-1})$; and (iii) the edge feature $\mathbf{e}_{i,j}^{l-1}$ that represents the relationship between $\mathbf{v}_i^{l-1}$ and $\mathbf{v}_j^{l-1}$ in $\mathcal{N}(\mathbf{v}_i^{l-1})$. Here, the message $\mathbf{m}$ that the vertex $\mathbf{v}_i^{l-1}$  received from its adjacent vertices $\mathcal{N}(\mathbf{v}^{l-1}_i)$ can be denoted as:
\begin{equation}
\begin{split}
& \mathbf{m}_{\mathcal{N}(\mathbf{v}^{l-1}_i)} = M(\mathbin\Vert ^{N}_{j=1} f(\mathbf{v}_j^{l-1},\mathbf{e}_{i,j}^{l-1})) \\
& f(\mathbf{v}_j^{l-1},\mathbf{e}_{i,j}^{l-1}) = 0 \quad \text{subject to} \quad \mathcal{A}^{l-1}_{i,j} = 0
\end{split}
\label{eq:message-passing}
\end{equation}
where $M$ is a differentiable function that aggregates messages produced from the adjacent vertices; $N$ denotes the number of vertices in the graph $\mathcal{G}^{l-1}$; $f(\mathbf{v}_j^{l-1},\mathbf{e}_{i,j}^{l-1})$ is a differentiable function which defines the influence of an $\mathbf{v}_i^{l-1}$'s adjacent vertex $\mathbf{v}_j^{l-1}$ on $\mathbf{v}_i^{l-1}$ through their edge $\mathbf{e}_{i,j}^{l-1}$; and $\mathbin\Vert $ is the aggregation operator to combine messages of all adjacent vertices of the $\mathbf{v}_i^{l-1}$. As a result, the vertex feature $\mathbf{v}^l_i$ can be produced as:
\begin{equation}
\begin{split}
   \mathbf{v}^l_i = G_v^l(\mathbf{v}^{l-1}_i,\mathbf{m}_{\mathcal{N}(\mathbf{v}^{l-1}_i)})
\end{split}
\label{eq:vertex}
\end{equation}
where $G_v^l$ denotes a differentiable function of the $l_{th}$ GNN layer, which updates each vertex feature for the graph $\mathcal{G}^{l}$.

Besides vertices, an edge feature $\mathbf{e}_{i,j}^l$ in the graph $\mathcal{G}^{l}$ can be either kept as the same to the its previous status $\mathbf{e}_{i,j}^{l-1}$ in the graph $\mathcal{G}^{l-1}$ \cite{kipf2016semi,hamilton2017inductive}  (denoted as GNN type 1), or updated during the propagation of GNNs \cite{gong2019exploiting,jo2021edge,isufi2021edgenets} (denoted as GNN type 2). Specifically, the edge feature $\mathbf{e}_{i,j}^l$ in $\mathcal{G}^{l}$ is computed based on: (i) its previous status $\mathbf{e}^{l-1}_{i,j}$ in $\mathcal{G}^{l-1}$; and (ii) the corresponding vertex features $\mathbf{v}^{l-1}_i$ and $\mathbf{v}^{l-1}_j$ in $\mathcal{G}^{l-1}$. Mathmatically speaking, the $\mathbf{e}^{l-1}_{i,j}$ can be computed as:
\begin{equation}
\begin{split}
    \mathbf{e}_{i,j}^l = 
    \begin{cases}
    G_e^l(\mathbf{e}^{l-1}_{i,j},g(\mathbf{v}^{l-1}_i, \mathbf{v}^{l-1}_j)) & \text{GNN type 2} \\
    \mathbf{e}^{l-1}_{i,j} &  \text{GNN type 1}
    \end{cases}
\end{split}
\end{equation}
where $G_e^l$ is a differentiable function of the $l_{th}$ GNN layer, which updates each edge feature for the graph $\mathcal{G}^{l}$, and $g$ is also a differentiable function that models relationship cues between $\mathbf{v}^{l-1}_i$ and $\mathbf{v}^{l-1}_j$. In summary, during the propagation of a GNN, the updating of vertex features and edge features are mutually influenced. Please refer to Hamilton et al. \cite{hamilton2020graph} and Dwivedi et al. \cite{dwivedi2020benchmarking} for more details.

\section{Problem formulation}
\label{sec:problem}

\textbf{Problem 1: manually defined task-agnostic graph topology.} With a manually defined graph topology (represented as an adjacency matrix) $\mathcal{A} \in \mathbb{R}^{|\mathcal{V}| \times |\mathcal{V}|}$ \cite{giles1998citeseer,yan2018spatial}, a pair of vertices $\mathbf{v}^{l}_i$ and $\mathbf{v}^{l}_j$ are connected when their relationship meets the pre-defined criteria $\mathcal{R}$, while no edge presented between a pair of vertices whose relationships are not considered by $\mathcal{R}$. Assuming $\mathcal{A}_{i,j} \in \mathcal{A}$ describes the presence of the edge $\mathbf{e}_{i,j}$ between vertices $\mathbf{v}_i$ and $\mathbf{v}_j$ in $\mathcal{G}$, then it is represented as
\begin{equation}
\begin{split}
    \mathcal{A}_{i,j} = 
    \begin{cases}
    1 & \{\mathbf{v}_i, \mathbf{v}_j \} \in \mathcal{R} \\
    0 &  \text{Otherwise}
    \end{cases}
\end{split}
\label{eq:edge}
\end{equation}
As described in Eqa. \ref{eq:message-passing}, the message passing of vertex features in the graph $\mathcal{G}$ depends on its adjacency matrix $\mathcal{A}$. Consequently, a manually defined adjacency matrix may not provide the task-specific message passing mechanism for the graph. In other words, properly exploring a task-specific adjacency matrix $\mathcal{\hat{A}} \in \mathbb{R}^{|\mathcal{V}| \times |\mathcal{V}|}$ for $\mathcal{G}$ would allow vertex messages to be passed via task-specific paths, and result in better graph analysis performances.


\noindent \textbf{Problem 2: message passing via single-value edges.} While edges are essential components for a graph and decide its message passing process, many existing approaches only use a single value as each edge's representation (denoted as $\mathbf{e}_{i,j} = [e_{i,j}(1)]$) to define either the edge's presence \cite{yan2018spatial,xie2020adversarial} or the strength of association between a pair of vertices \cite{isufi2021edgenets,song2021uncertain} (i.e., $e_{i,j}(1) = \mathcal{A}_{i,j}$). Let's define each vertex in a single-value edge feature-based graph $\mathcal{G}$ as $\mathbf{v}_j = [v_j(1), v_j(2), \cdots, v_j(K)]$ ($j = 1, 2, \cdots, N$). Then, the function $f$ in Eqa. \ref{eq:message-passing} can be re-written as: 
\begin{equation}
\begin{split}
f(\mathbf{v}_j,\mathbf{e}_{i,j}) = f([v_j(1),
\cdots, v_j(K)], [e_{i,j}(1)])
\end{split}
\label{eq:single_edge}
\end{equation}
where the impact of the vertex $\mathbf{v}_j$ on its adjacent vertex $\mathbf{v}_i$ is only controlled by a single value $e_{i,j}(1)$, which may fail to include all crucial relationship cues between vertices $\mathbf{v}_i$ and $\mathbf{v}_j$. Consequently, the messages passed via such single-value edges may not be optimal.

\section{Methodology}

\noindent In this paper, we propose a novel graph representation learning framework (GRATIS) that produces a graph representation with a task-specific topology and multi-dimensional edge features to describe an arbitrary graph or non-graph data (e.g., pre-defined graphs, images, text data, etc.). Specifically, our approach takes an arbitrary input $\mathcal{D}^{\text{in}}$, and produces a task-specific graph representation $\mathcal{\hat{G}}(\mathcal{V}, \mathcal{\hat{E}})$ ($\mathcal{\hat{G}}(\mathcal{\hat{V}}, \mathcal{\hat{E}})$ for non-graph data) which consists of $N$ vertices $\mathcal{V} \subseteq \{\mathbf{v}_i \in \mathbb{R}^{1 \times K} \}$ ($i = 1, 2, \cdots, N$), and a set of edges whose presences are defined by a binary adjacency matrix $\mathcal{\hat{A}} \in \mathbb{R}^{N \times N}$, where each of these presented edges is described by a task-specific multi-dimensional edge feature. These edges can be denoted as $\mathcal{\hat{E}} \subseteq \{ \mathbf{\hat{e}}_{i,j} = \mathbf{\hat{e}}(\mathbf{v}_i, \mathbf{v}_j) \mid \mathbf{v}_i, \mathbf{v}_j \in \mathcal{V} \quad \text{and} \quad \mathcal{\hat{A}}_{i,j} = 1 \}$.

The proposed framework consists of three modules: (i) the \textbf{Graph Definition (GD)} that produces a basic graph representation $\mathcal{G}^{\text{B}}(\mathcal{V}, \mathcal{E})$ from the input data $\mathcal{D}^{\text{in}}$. The $\mathcal{G}^{\text{B}}(\mathcal{V}, \mathcal{E})$ is defined by a set of vertex features $\mathcal{V}$, a basic topology (adjacency matrix) $\mathcal{A}$, and a set of basic edge features $\mathcal{E} \subseteq \{ \mathbf{e}_{i,j} = \mathbf{e}(\mathbf{v}_i, \mathbf{v}_j) \mid \mathbf{v}_i, \mathbf{v}_j \in \mathcal{V} \quad \text{and} \quad \mathcal{A}_{i,j} = 1 \}$; (ii) the \textbf{Task-specific Topology Prediction (TTP)} that produces a task-specific graph topology, i.e., replacing the basic graph topology defined by $\mathcal{A}$ with a task-specific adjacency matrix $\mathcal{\hat{A}} \in \mathbb{R}^{N \times N}$; and (iii) the \textbf{Multi-dimensional Edge Feature Generation (MEFG)} that specifically assigns a task-specific multi-dimensional edge feature $\mathbf{\hat{e}}_{i,j}$ to each presented edge ($\mathcal{\hat{A}}_{i,j} = 1$), describing multiple task-specific relationship cues between vertices $\mathbf{v}_i$ and $\mathbf{v}_j$ (i.e., replacing basic edge features $\mathcal{E}$ with task-specific multi-dimensional edge features $\mathcal{\hat{E}}$). As a result, a task-specific graph representation $\mathcal{\hat{G}}(\mathcal{V}, \mathcal{\hat{E}})$ (or $\mathcal{\hat{G}}(\hat{\mathcal{V}}, \mathcal{\hat{E}})$ for non-graph data) whose topology is defined by $\mathcal{\hat{A}}$ can be obtained from any arbitrary input $\mathcal{D}^{\text{in}}$ (i.e., if the input non-graph data is represented by a set of vectors, we concatenate them as a single matrix $\mathcal{D}^{\text{in}}$). The pseudocode for the entire framework of the GRATIS is demonstrated in Algorithm 1.

\begin{algorithm}[t]
\label{alg:alg-vertex-encoding}
  \caption{Task-specific Graph Representation Generation}  
  \begin{algorithmic}[1]
    \Require 
        A input graph/non-graph data $\mathcal{D}^{\text{in}}$; an 'Backbone'; an GD module; an TTP module; and an MEFG module.
    \Ensure  
       A task-specific graph representation $\mathcal{\hat{G}}(\mathcal{\hat{V}}, \mathcal{\hat{E}})$.

      \State Generating global contextual representation $ X  \leftarrow \text{Backbone}(\mathcal{D}^{\text{in}})$

      \State Generating basic graph representation $ \mathcal{G}^{\text{B}}(\mathcal{V}, \mathcal{E}))  \leftarrow \text{GD}(\mathcal{D}^{\text{in}}$
      
      \State Generating the task-specific topology and vertex features $ [\mathcal{\hat{V}},\mathcal{\hat{A}}] \leftarrow \text{TTP}(\mathcal{V}, \mathcal{A}, X)$ 
      
      \State Generating task-specific multi-dimensional edge features $ \mathcal{\hat{E}} \leftarrow \text{MEFG}(\mathcal{\hat{V}}, \mathcal{\hat{A}}, X)$      
      
      \State Combining $\mathcal{\hat{V}}$ with the generated $\mathcal{\hat{E}}$ and $\mathcal{\hat{A}}$ to construct the final task-specific graph representation $\mathcal{\hat{G}}(\mathcal{\hat{V}}, \mathcal{\hat{E}})$

  \end{algorithmic}  
\end{algorithm}

\begin{figure}[htb]
  \centering
  \includegraphics[width=1\columnwidth]{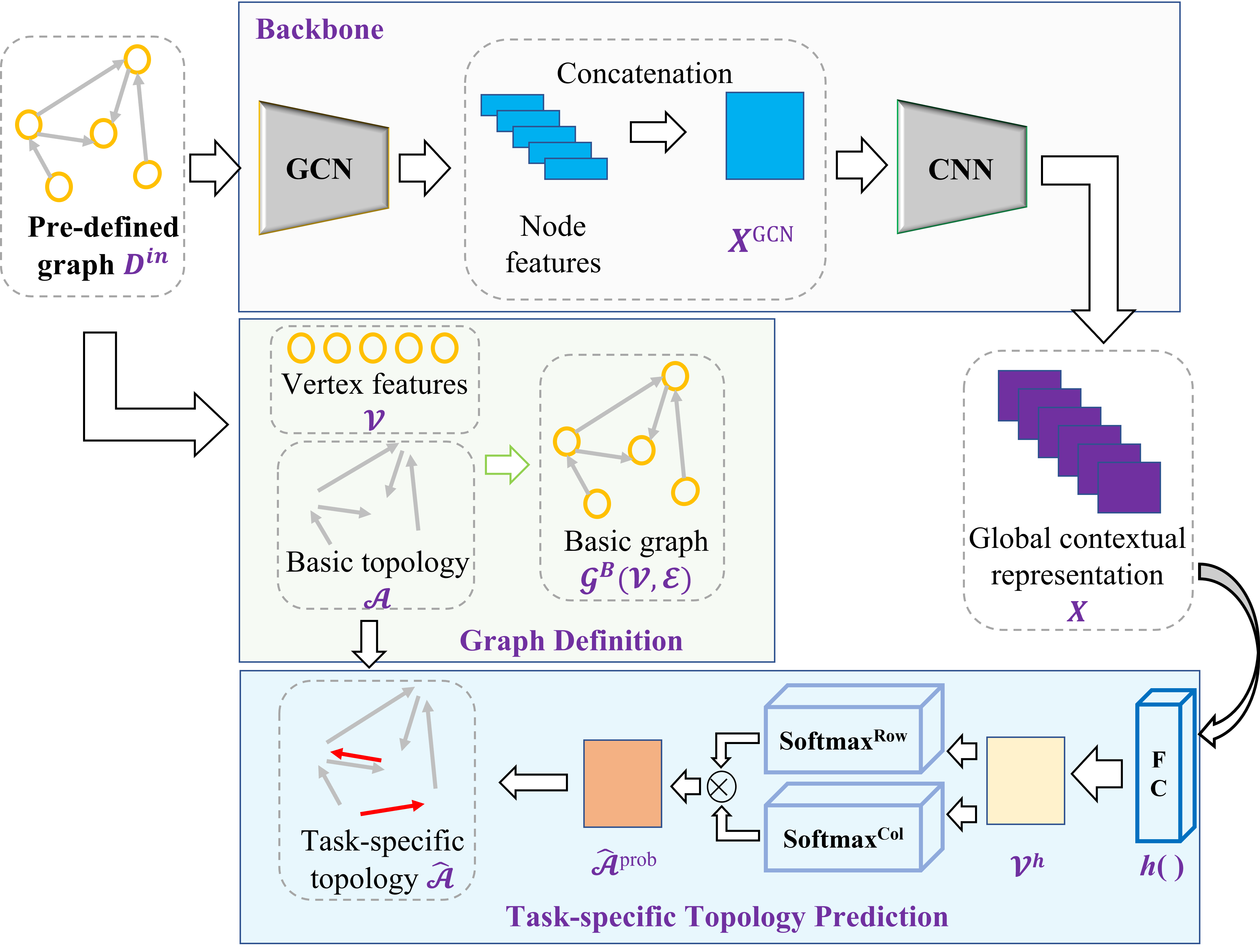}
  \caption{Illustration of the GCN-CNN Backbone, Graph Definition (GD) and the Task-specific Topology Prediction (TTP) modules for \textbf{graph data}. The GCN in the backbone first produces a set of vertex features from the input graph $\mathcal{D}^{\text{in}}$, which are concatenated as a matrix $X^{\text{GCN}}$. Then, the CNN yields the global contextual representation $X$ from the $X^{\text{GCN}}$. The GD module simply treated the original input graph as the basic graph. The TTP module first produces an adjacency probability matrix $\mathcal{\hat{A}}^{\text{prob}}$ from $X$, which are then combined with the basic graph topology to generate the final task-specific adjacency matrix $\mathcal{\hat{A}}$.}
\label{fig:TTP_graph}
\end{figure}

As shown in Fig. \ref{fig:method_overview}, the $\mathcal{D}^{\text{in}}$ is firstly fed to a backbone which can be any suitable machine learning model for $\mathcal{D}^{\text{in}}$, e.g., a GNN for graph data or a CNN/Transformer for non-graph data. Then, a global contextual representation $X$ can be obtained. We formulated this process as:
\begin{equation}
  X = \text{Backbone}(\mathcal{D}^{\text{in}})
\end{equation}
In this paper, a CNN or a transformer is employed as the backbone to directly extract $X$ from non-graph data, where the $X$ is a set of latent feature maps whose sizes depend on the input data $\mathcal{D}^{\text{in}}$. Meanwhile, we propose a GCN-CNN network as the backbone to project the input graph data $\mathcal{D}^{\text{in}}$ to an $X \in \mathbb{R}^{N \times N \times D}$, where $N$ denotes the number of vertices, i.e. the backbone projects the input graph to a $N \times N \times D$ dimensional latent space, summarising its global contextual information (illustrated in Fig. \ref{fig:TTP_graph}). Specifically, the GCN part first projects the input graph $\mathcal{D}^{\text{in}}$ to a matrix $X^{\text{GCN}}$ with the size of $N \times K$, where $K$ is the original dimensionality of the vertex for the input graph $\mathcal{D}^{\text{in}}$, which is defined as
\begin{equation}
    \begin{split}
        X^{\text{GCN}} &= \text{GCN}(\mathcal{D}^{\text{in}}) \\
        \mathcal{D}^{\text{in}} &= \mathcal{G}^{\text{in}}(\mathcal{V}^{\text{in}}, \mathcal{E}^{\text{in}}) \\
    \end{split}
\end{equation}
where $\mathcal{V}^{\text{in}} \subseteq \{\mathbf{v}_i^{\text{in}} \in \mathbb{R}^{1 \times K} \mid i = 1,2 \cdots N  \}$, and $\mathcal{E}^{\text{in}} \subseteq \{ \mathbf{e}_{i,j}^{\text{in}} = \mathbf{e}(\mathbf{v}_i, \mathbf{v}_j) \mid \mathbf{v}_i^{\text{in}}, \mathbf{v}_j^{\text{in}} \in \mathcal{V}^{\text{in}} \}$ are the vertices set and edges set of the original input graph. Then, the CNN part produces the global contextual representation $X \in \mathbb{R}^{N\times N\times D}$ from the $X^{\text{GCN}}$: 
\begin{equation}
\begin{split}
    &X = \mathbf{\bar{M}}_1 \mathbf{M}_2^\mathbf{T} \\
    &\mathbf{M}_1 = X^{\text{GCN}}W_1 \\
    &\mathbf{M}_2 = X^{\text{GCN}}W_2 \\
\end{split}
\label{eq:topology_global}
\end{equation}
where $W_1 \in \mathbb{R}^{K\times D^2}$ and $W_2 \in \mathbb{R}^{K\times D}$ are learnable weight matrices. During this process, the the $W_1$ projects $X^{\text{GCN}}$ to a high-dimensional matrix $\mathbf{M}_1  \in \mathbb{R}^{N\times D^2}$ (i.e., $\mathbf{M}_1$ has $N$ rows and $D^2$ columns), while $W_2$ projecting each row vector of $X^{\text{GCN}}$ from $K$ dimension to $D$ dimension, resulting in a matrix $\mathbf{M}_2 \in \mathbb{R}^{N\times D}$. Then, the $\mathbf{M}_1$ is reshaped as $\mathbf{\bar{M}}_1 \in \mathbb{R}^{ND\times D}$, and we conduct matrix multiplication between $\mathbf{\bar{M}}_1$ and $\mathbf{M}_2$ and reshape the obtained matrix to achieve a global contextual representation $X \in \mathbb{R}^{N\times N\times D}$ that summarizes entire contextual information of the $\mathcal{D}^{\text{in}}$.

\begin{figure*}[htb]
  \centering
  \includegraphics[width=1.8\columnwidth]{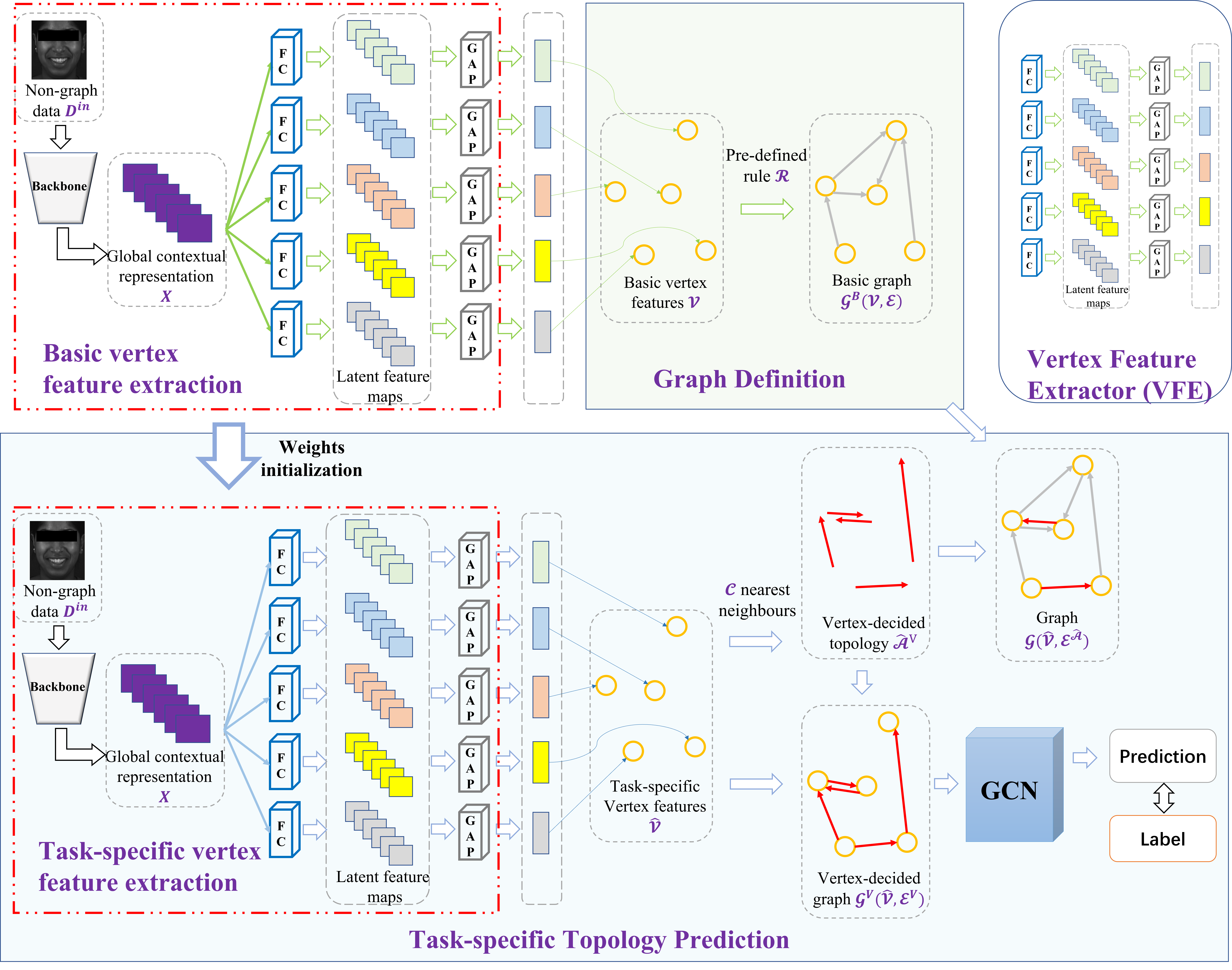}
  \caption{Illustration of the Graph Definition (GD) and the Task-specific Topology Prediction (TTP) modules for \textbf{non-graph data}. In the GD module, the VFE is trained with a standard MLP-based predictor to provide a set of basic vertex features, based on which the basic graph is defined. Then, the TTP module re-trains the pre-trained VFE with a GCN, allowing a $\hat{\text{VFE}}$ to be learned to produce task-specific vertex features and a vertex-decided graph topology, which are combined with the basic graph topology to generate the graph $\mathcal{G}^{\text{V}}(\mathcal{\hat{V}},\mathcal{E}^{V})$.}
\label{fig:TTP_nongraph}
\end{figure*}



\subsection{Graph Definition}
\label{subsec:approach-graph definition}

\noindent Given the input data $\mathcal{D}^{\text{in}}$, we first categorize it as either a pre-defined graph $\mathcal{D}^{\text{in}} = \mathcal{G}^{\text{in}}(\mathcal{V}^{\text{in}}, \mathcal{E}^{\text{in}})$ or a non-graph data $\mathcal{D}^{\text{in}}$ (e.g., image, text, etc.), and then obtain the vertex features $\mathcal{V}$, basic topology $\mathcal{A}$ and basic edge features $\mathcal{E}$ to build its basic graph representation $\mathcal{G}^{\text{B}}(\mathcal{V}, \mathcal{E})$:
\begin{itemize}

    \item \textbf{Pre-defined graph:} For a pre-defined single-value edge graph $\mathcal{G}^{\text{in}}(\mathcal{V}^{\text{in}}, \mathcal{E}^{\text{in}})$ that has an adjacency matrix $\mathcal{A}^{\text{in}}$, we directly employ its original vertices $\mathcal{V}^{\text{in}}$, topology $\mathcal{A}^{\text{in}}$ and edge features $\mathcal{E}^{\text{in}}$ to define vertex features, basic topology, and basic edge features for $\mathcal{G}^{\text{B}}$. This can be formulated as:
    \begin{equation}
        \mathcal{V} = \mathcal{V}^{\text{in}}, \quad
        \mathcal{A} =  \mathcal{A}^{\text{in}}, \quad
        \mathcal{E} = \mathcal{E}^{\text{in}} 
    \end{equation}

    \item \textbf{Non-graph data:} For a non-graph data, we further  propose a Vertex Feature Extraction (VFE) module that is made up of $N$ vertex feature extractors, each of which consists of a fully connected layer (FC) and a global average pooling (GAP) layer. As shown in Fig. \ref{fig:TTP_nongraph}, the $i_{th}$ vertex feature extractor takes the global contextual representation $X$ as the input, and produces a set of latent feature maps which are then further processed by the GAP layer as a vector of $K$ dimensions. This vector represents the feature of the $i_{th}$ vertex $\mathbf{v}_i$ in the basic graph $\mathcal{G}^{\text{B}}$. As a result, the vertices set $\mathcal{V}$ contains $N$ vectors (each has $K$ dimensions), representing $N$ vertices and their features. In addition, \textbf{if the non-graph data is represented by a set of vectors/multi-channel data}, we directly treat each vector as a vertex, and the matrix that concatenates all vectors as the global contextual representation $X$. After that, we manually define edge presence (the basic adjacency matrix $\mathcal{A}$) in the basic graph $\mathcal{G}^{\text{B}}$ according to a human interpretable rule $\mathcal{R}$ (e.g., Euclidean distance or correlation between vertex features) depending on the data $\mathcal{D}^{\text{in}}$ and the task, where the basic edge representation between each pair of connected vertices is defined as $1$. These can be formulated as
    \begin{equation}
    \begin{split}
        &\mathcal{V} = \text{VFE}(X) \subseteq \{\mathbf{v}_i \in \mathbb{R}^{1 \times K}  \mid i = 1, 2, \cdots, N \} \\
        &\mathcal{E} \subseteq \{ \mathbf{e}_{i,j} =  1 \mid \mathbf{v}_i, \mathbf{v}_j \in \mathcal{V} \quad \text{and} \quad \mathcal{A}_{i,j} = 1 \}
    \end{split}
    \label{eq:gd-vertex}
    \end{equation}
    which is subject to:
    \begin{equation}
    \begin{split}
        \mathcal{A}_{i,j} = 
        \begin{cases}
        1 & \{\mathbf{v}_i, \mathbf{v}_j \} \in \mathcal{R} \\
        0 &  \text{Otherwise}
        \end{cases}
    \end{split}
    \label{eq:raw_a_nongraph}
    \end{equation}

\end{itemize}

\subsection{Task-specific Topology Prediction}
\label{subsec:approach-topology}

\noindent This section introduces our task-specific Topology Prediction (TTP) module which aims to address the \textbf{Problem 1} defined in Sec. \ref{sec:problem}. This module takes the produced global contextual representation $X$, the basic adjacency matrix $\mathcal{A}$ and vertex features $\mathcal{V}$ as the input, and generates a task-specific adjacency matrix $\mathcal{\hat{A}} \in \mathbb{R}^{N \times N}$ to describe the topology of the graph $\mathcal{G}$.It also further generates a set of task-specific vertex features for the graph representation of the non-graph data, which can be formulated as:
\begin{equation}
   [\mathcal{\hat{V}},\mathcal{\hat{A}})] = \text{TTP}(X, \mathcal{A}, \mathcal{V})
\end{equation}
We again propose different TTP implementations for pre-defined graph and non-graph data, respectively, which are explained in the following:
\begin{itemize}

    \item \textbf{Pre-defined graph:} Building upon the global contextual representation $X$, a trainable linear function $h$ (a FC layer) is firstly introduced to project $X$ to a matrix $X^h = h(X)$ that has the same size as the target adjacency matrix $\mathcal{\hat{A}}$ ($N \times N$ dimensions). Then, the TTP individually conducts Softmax operations on $X^h$ along its row and column vectors, and combines two generated matrices via element-wise product $\otimes$. Consequently, an adjacency probability matrix $\mathcal{\hat{A}}^{\text{prob}}$ can be obtained by:
    \begin{equation}
      \mathcal{\hat{A}}^{\text{prob}} = (\text{Softmax}^{\text{Row}}(h(X)) \otimes \text{Softmax}^{\text{Column}}(h(X))
    \end{equation}    
    where each component $\mathcal{\hat{A}}^{\text{prob}}_{i,j}$ in the $\mathcal{\hat{A}}^{\text{prob}}$ ranges from 0 to 1. As a result, the presence probability of an edge $\mathbf{e}_{i,j}$ (denoted as the $\mathcal{\hat{A}}^{\text{prob}}_{i,j}$) is obtained by taking the global context $X$ into consideration. Meanwhile, the vertex features $\mathcal{\hat{V}} = \mathcal{V}$ is defined for graph data (i.e., the $\mathcal{\hat{V}}$ is defined as the same to the basic vertex features $\mathcal{V}$). This process is also illustrated in Fig. \ref{fig:TTP_graph}.

    \item \textbf{Non-graph data:} Different from graph data whose vertex features are pre-defined and fixed, vertex features of non-graph data are dependant on the learning process of the VFE. We propose to train the VFE as the $\hat{\text{VFE}}(X)$ which additionally encodes task-specific associations among vertices into vertex features, i.e., the learned vertex features encode not only task-specific object representations contained in the input data but also the association among them. In other words, the presence of each edge is decided by the corresponding pair of vertices using a specific rule (e.g., distances or similarity between vertices), where the edge feature of each presented edge is $1$. This process can be denoted as:
    \begin{equation}
    \begin{split}
        &\mathcal{\hat{V}} = \hat{\text{VFE}}(X) \subseteq \{\mathbf{\hat{v}}_i \in \mathbb{R}^{1 \times K}  \mid i = 1, 2, \cdots, N \} \\
        &\mathcal{E}^{\mathcal{\hat{V}}} \subseteq \{ \mathbf{e}^{\text{V}}_{i,j} =  1 \mid \mathbf{\hat{v}}_i, \mathbf{\hat{v}}_j \in \mathcal{\hat{V}} \quad \text{and} \quad \mathcal{\hat{A}}^{\text{V}}_{i,j} = 1 \}
    \end{split}
    \end{equation}
    which is conditioned on:
    \begin{equation}
    \begin{split}
        \mathcal{\hat{A}}^{\text{V}}_{i,j} = 
        \begin{cases}
        1 & \mathbf{\hat{v}}_j \in \mathcal{C}(\mathbf{v}_i) \\
        0 &  \text{Otherwise}
        \end{cases}
    \end{split}
    \label{eq:C-nearest}
    \end{equation}     
    where $\mathcal{C}(\mathbf{v}_i)$ denotes the $C$ nearest neighbour vertices of the vertex $\mathbf{v}_i$. In this paper, the vertex-decided adjacency matrix is obtained by connecting each vertex to its $C$ nearest neighbour vertices. To achieve the $\hat{\text{VFE}}$, we further attach a GCN predictor to produce a prediction from the obtained graph $\mathcal{G}^{\text{V}}(\mathcal{\hat{V}},\mathcal{E}^{V})$, and use the loss between the prediction and labels to supervise the training of the $\hat{\text{VFE}}$ (visualized in Fig. \ref{fig:TTP_nongraph}). This way, the $\hat{\text{VFE}}$ learns to directly produce both task-specific vertex features $\mathcal{\hat{V}}$ and a task-specific vertex-decided adjacency matrix $\mathcal{\hat{A}}^{\text{V}}$ for the input non-graph data, where the $\mathcal{\hat{A}}^{\text{V}}$ allows the graph $\mathcal{G}^{\text{V}}(\mathcal{\hat{V}},\mathcal{E}^{V})$ to have optimal connectivity/message passing paths among vertices.
\end{itemize}

After obtaining a task-specific probability adjacency matrix $\mathcal{\hat{A}}^{\text{prob}}$ for a pre-defined graph $\mathcal{G}$ or a task-specific vertex-decided adjacency matrix $\mathcal{\hat{A}}^{\text{V}}$ for the graph representation $\mathcal{G}$ of non-graph data, we set two rules to determine the presence of each edge in the task-specific graph representation $\mathcal{\hat{G}}$ (i.e., the edge presence as defined in the task-specific adjacency matrix $\mathcal{\hat{A}}$): (i) each edge's presence probability encoded in $\mathcal{\hat{A}}^{\text{prob}}$ or the binary presence status encoded in $\mathcal{\hat{A}}^{\text{V}}$; and (ii) each edge's presence status defined in the basic adjacency matrix $\mathcal{A}$, as we hypothesize that edges defined in $\mathcal{A}$ may contain crucial relationship cues reflecting the underlying relationship among its vertices, which may be disregarded during the generation of $\mathcal{\hat{A}}^{\text{prob}}$ or $\mathcal{\hat{A}}^{\text{V}}$. Specifically, for a pre-defined graph, the $\mathcal{\hat{A}}$ is obtained by: 
\begin{equation}
\mathcal{\hat{A}}_{i,j} = 
\begin{cases}
1 & \mathcal{\hat{A}}^{\text{prob}}_{i,j} \geqslant \theta \quad  \text{or} \quad  \mathcal{A}_{i,j} = 1 \\
0 &  \text{Otherwise}
\end{cases}
\label{eq:ttp_graph}
\end{equation}    
where $\theta \in [0,1]$ is a threshold. Meanwhile, for graph representations of non-graph data, $\mathcal{\hat{A}}$ can be formulated as: 
\begin{equation}
\mathcal{\hat{A}}_{i,j} = 
\begin{cases}
1 &   \mathcal{\hat{A}}^{\text{V}} = 1 \quad  \text{or} \quad  \mathcal{A}_{i,j} = 1 \\
0 &  \text{Otherwise}
\end{cases}
\label{eq:ttp_nongraph}
\end{equation}
For both equations listed above, the term $\mathcal{\hat{A}}^{\text{prob}}_{i,j} \geqslant \theta \quad  \text{or} \quad  \mathcal{A}_{i,j} = 1$ in Eqa. \ref{eq:ttp_graph} and the term $\mathcal{\hat{A}}^{\text{V}} = 1 \quad  \text{or} \quad  \mathcal{A}_{i,j} = 1$ in Eqa. \ref{eq:ttp_nongraph} can be re-written as the $\text{TTP}(X, \mathcal{A})$. As a result, \textbf{the Problem 1 defined in Sec. \ref{sec:problem} is addressed} by replacing the manually defined rule $\mathcal{R}$ with the task-specific rule $\text{TTP}(X, \mathcal{A})$, i.e., the basic graph $\mathcal{G}^{\text{B}}(\mathcal{V}, \mathcal{E})$ is updated as the graph $\mathcal{G}(\mathcal{V}, \mathcal{E}^{\hat{A}})$ that has task-specific topology $\mathcal{E}^{\hat{A}} \subseteq \{ \mathbf{e}^{\hat{A}}_{i,j} =  1 \mid \mathbf{v}_i, \mathbf{v}_j \in \mathcal{V} \quad \text{and} \quad \mathcal{\hat{A}}_{i,j} = 1 \}$. As a result, the Eqa. \ref{eq:edge} is re-written as:
\begin{equation}
\begin{split}
\mathcal{\hat{A}}_{i,j} = 
\begin{cases}
1 & \{\mathbf{v}_i, \mathbf{v}_j \} \in \text{TTP}(X, \mathcal{A}) \\
0 &  \text{Otherwise}
\end{cases}
\end{split}
\label{eq:gtp-edge}
\end{equation}
In summary, during the generation of tasks-specific graph topology $\mathcal{\hat{A}}$, each edge's presence $\mathcal{\hat{A}}_{i,j}$ is decided by not only the corresponding vertices $\mathbf{v}_i$ and $\mathbf{v}_j$, but also the global contextual information contained in $X$. This further leads the $\mathcal{\hat{A}}_{i,j}$ to be globally optimal, i.e., the graph has a globally optimal and task-specific message passing paths.

\subsection{Multi-dimensional Edge Feature Generation}
\label{subsec:approach-multi-dimensional edge feature}


\noindent Once all vertex features $\mathcal{\hat{V}}$ and the task-specific topology $\mathcal{\hat{A}}$ are obtained, we propose a Multi-dimensional Edge Feature Generation (MEFG) module that further learns multiple task-specific relationship cues between vertices to describe each presented edge in a $1 \times K$ dimensional space, i.e., assigning each presented edge with a multi-dimensional feature $\mathbf{\hat{e}}_{i,j} \in \mathbb{R}^{1 \times K}$, describing the optimal message passing mechanism within the graph. This module is trained under the supervision of the target graph analysis task. The hypothesis of the MEFG module is that the relationship cues between a pair of connected multi-dimensional vertices $\mathbf{v}_i$ and $\mathbf{v}_j$ can not be optimally described by a single dimensional edge feature. Moreover, such relationship cues are not only contained in their vertex features but also reflected by the global context of the graph. For example, if we treat local facial regions as vertices (e.g., mouth and eyes), the muscle movements of mouth and eyes sometimes also lead to muscle movements in other facial regions (e.g.,the nose and cheeks).

\begin{figure}[htb]
  \centering
  \includegraphics[height=0.8\columnwidth]{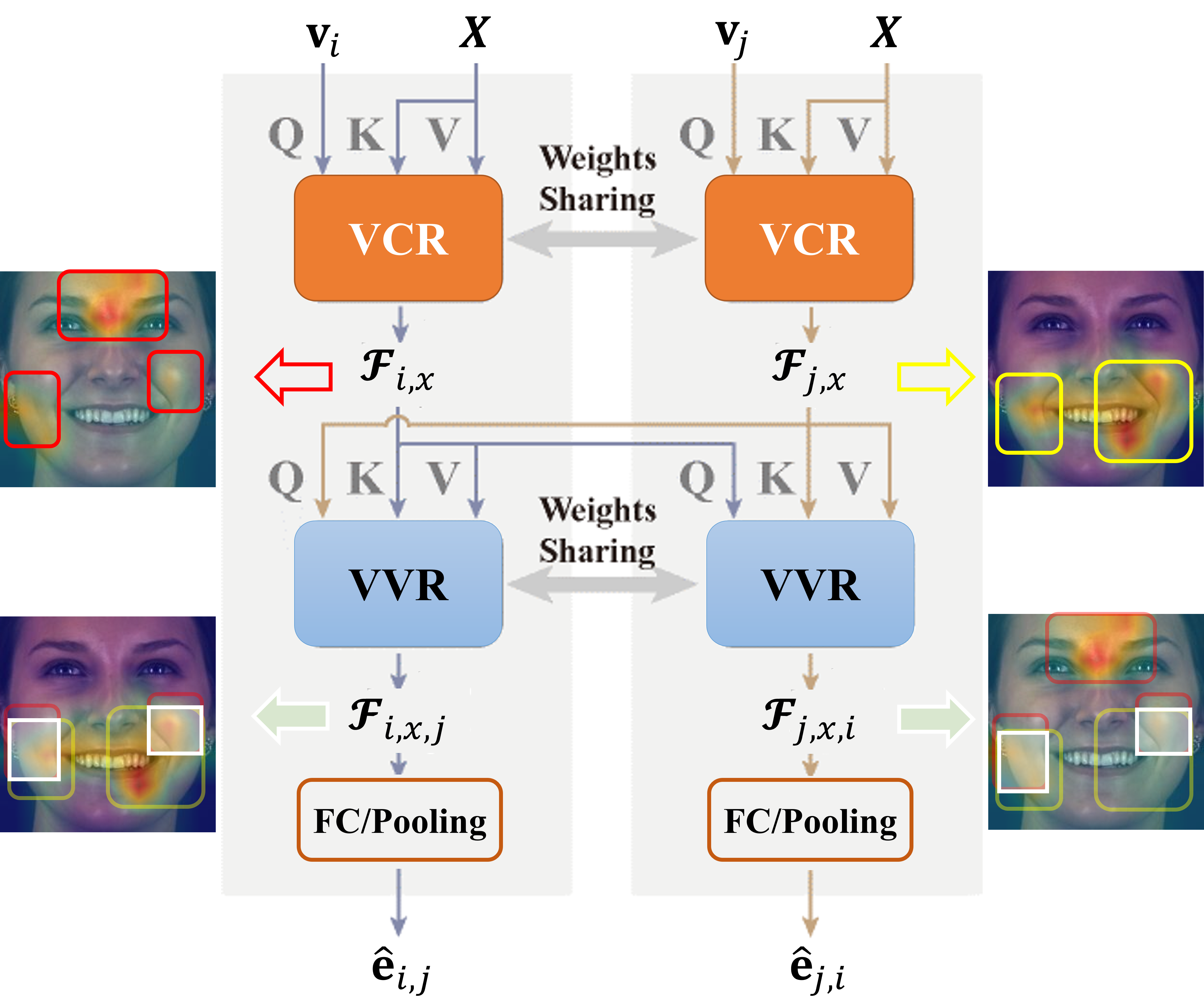}
  \caption{Illustration of the proposed MEFG module, where we visualize the outputs of VCR and VVR in an example face image. The \textbf{VCR} module first individually locates cues that are related to $i_{th}$ and $j_{th}$ vertices from the global contextual representation $\textbf{X}$, producing vertex-context relationship features $\mathcal{F}_{i,x}$ and $\mathcal{F}_{j,x}$ (visualized as the activated face areas depicted in red and yellow cubes). Then, the \textbf{VVR} module further extracts cues $\mathcal{F}_{i,x,j}$ and $\mathcal{F}_{j,x,i}$ in the global contextual representation $\textbf{X}$, which are related to both $i_{th}$ and $j_{th}$ vertices. Finally, a FC or Pooling layer is introduced to generate final task-specific multi-dimensional edge features $\mathbf{\hat{e}}_{i,j}$ and $\mathbf{\hat{e}}_{j,i}$.}
\label{fig:MEFG}
\end{figure}

In this sense, we propose to generate a task-specific multi-dimensional edge feature $\mathbf{\hat{e}}_{i,j}$ for each presented edge ($\mathcal{\hat{A}}_{i,j} = 1$) by considering not only its corresponding vertex features $\mathbf{v}_i$ and $\mathbf{v}_j$ but also the global contextual representation $X$, which can be formulated as:
\begin{equation}
\begin{split}
    &\mathbf{\hat{e}}_{i,j} = \text{MEFG}(X, \mathbf{v}_i, \mathbf{v}_j) \\
    &\mathcal{\hat{E}} \subseteq \{ \mathbf{\hat{e}}_{i,j} = \mathbf{\hat{e}}(\mathbf{v}_i, \mathbf{v}_j)\mid \mathbf{v}_i, \mathbf{v}_j \in \mathcal{V} \quad \text{and} \quad \mathcal{\hat{A}}_{i,j} = 1 \}    
\end{split}
\end{equation}
Specifically, the MEFG module consists of two blocks: a \textbf{vertex-context relationship modelling (VCR)} block that first locates each vertex-related cues in the global contextual representation $X$, and a \textbf{vertex-vertex relationship modelling (VVR)} block which further extracts context-aware vertex-vertex relationship features from the produced vertex-context representation (i.e., the outputs of the VCR) to generate final multi-dimensional edge features. The full process is also illustrated in Fig. \ref{fig:MEFG}.

\textbf{VCR:} To learn multiple task-specific relationship cues between $\mathbf{v}_i$ and $\mathbf{v}_j$ and use them as the multi-dimensional edge feature $\mathbf{e}_{i,j}$ or $\mathbf{e}_{j,i}$, the VCR block takes vertex features $\mathbf{v}_i$ and $\mathbf{v}_j$ (for non-graph data, it takes the latent feature maps corresponding to $\mathbf{v}_i$ and $\mathbf{v}_j$, which are also denoted as $\mathbf{v}_i$ and $\mathbf{v}_j$ in the following contents) and the global contextual representation $X$ as input. It first conducts cross attention between $\mathbf{v}_i$ and $X$ as well as $\mathbf{v}_j$ and $X$. Here, the vertex features/latent feature maps $\mathbf{v}_i$ and $\mathbf{v}_j$ are independently used as queries to locate vertex-context relationship features $\mathcal{F}_{i,x}$ and $\mathcal{F}_{j,x}$ in $X$ (i.e., $X$ is treated as the key and value for attention operations). Mathematically speaking, this process can be represented as:
\begin{equation}
\begin{split}
    \mathcal{F}_{i,x} = \text{VCR}(\mathbf{v}_i, X) \\
    \mathcal{F}_{j,x} = \text{VCR}(\mathbf{v}_j, X)    
\end{split}
\end{equation}
with the cross attention operation in VCR defined as:
\begin{equation}
     \text{VCR}(A, B) = \text{softmax}(\frac{A W_q (B W_k)^T}{\sqrt{d_k} }) B W_v
\label{eq:VCR}
\end{equation}
where $W_q$, $W_k$ and $W_v$ are learnable weight vectors or matrices (depending on the shape of the input data) for the query, key and value encoding, respectively, and $d_k$ is a scaling factor set to the same as the number of the $B$'s channels. Subsequently, the produced $\mathcal{F}_{i,x}$ and $\mathcal{F}_{j,x}$ contain the vertex $\mathbf{v}_i$-related and vertex $\mathbf{v}_j$-related task-specific cues extracted from the global contextual representation $X$.

\textbf{VVR:} Based on the $\mathcal{F}_{i,x}$ and $\mathcal{F}_{j,x}$, the VVR block further extracts task-specific context cues that relate to both vertices. VVR is also a cross-attention block that has the same form as VCR (Eq.~\ref{eq:VCR}). In particular, it individually takes $\mathcal{F}_{i,x}$ as the query and $\mathcal{F}_{j,x}$ as the key and value, as well as $\mathcal{F}_{j,x}$ as the query and $\mathcal{F}_{i,x}$ as the key and value, producing two context-aware vertex-vertex relationship features $\mathcal{F}_{i,x,j}$ and $\mathcal{F}_{j,x,i}$, respectively. Here, the $\mathcal{F}_{i,x,j}$ encodes $\mathcal{F}_{i,x}$-related cues in the $\mathcal{F}_{j,x}$, while the $\mathcal{F}_{j,x,i}$ encoding $\mathcal{F}_{j,x}$-related cues in the $\mathcal{F}_{i,x}$. In other words, the context-aware vertex-vertex relationship features $\mathcal{F}_{i,x,j}$ and $\mathcal{F}_{j,x,i}$ contain cues that not only come from the whole context, but also relate to both vertex $\mathbf{v}_i$ and $\mathbf{v}_j$. We formulated this process as:
\begin{equation}
\begin{split}
\mathcal{F}_{i,x,j} = \text{VVR}(\mathcal{F}_{i,x}, \mathcal{F}_{j,x}) \\
\mathcal{F}_{j,x,i} = \text{VVR}(\mathcal{F}_{j,x}, \mathcal{F}_{i,x})    
\end{split}
\end{equation}
Depending on the data shape, we finally employ either a pooling layer or a fully-connected layer, to flatten $\mathcal{F}_{i,x,j}$ and $\mathcal{F}_{j,x,i}$ to a pair of multi-dimensional edge feature vectors $\mathbf{e}_{i,j}$ and $\mathbf{e}_{j,i}$ (we denote this operation as $L$):
\begin{equation}
\begin{split}
\mathbf{\hat{e}}_{i,j} = L(\mathcal{F}_{i,x,j}) \\
\mathbf{\hat{e}}_{j,i} = L(\mathcal{F}_{j,x,i})
\end{split}
\end{equation}
As a result, each of the produced multi-dimensional edge feature encodes task-specific cues from the whole contextual cues of the input data $\mathcal{D}^{\text{in}}$, which relate to both vertex $\mathbf{v}_i$ and $\mathbf{v}_j$.

This way, each edge feature can be represented as $\mathbf{e}_{i,j} = [e_{i,j}(1), e_{i,j}(2), \dots, e_{i,j}(K)]$. Subsequently, the proposed MEFG module \textbf{addresses the Problem 2} by passing messages via multi-dimensional task-specific edge features within the graph, i.e., the Eqa. \ref{eq:single_edge} is re-written as
\begin{equation}
f(\mathbf{v}_j,\mathbf{\hat{e}}_{i,j})  =   f([v_j(1), \cdots, v_j(K)], [\hat{e}_{i,j}(1), \cdots, \hat{e}_{i,j}(K)])
\end{equation}
where the impact of the vertex $\mathbf{v}_j$ on its adjacent vertex $\mathbf{v}_i$ is jointly controlled by $K$ task-specific relationship cues $[\hat{e}_{i,j}(1), \cdots, \hat{e}_{i,j}(K)]$. In short, the final produced task-specific multi-dimensional edge feature-based graph can be represented as $\mathcal{\hat{G}}(\mathcal{V}, \mathcal{\hat{E}})$, whose message passing mechanism is defined by the globally optimal and task-specific topology $\mathcal{\hat{A}}$ and multi-dimensional edge features $\mathcal{\hat{E}}$.

\section{Experiments}
\label{sec:experiments}

\noindent To evaluate the effectiveness of the proposed graph representation learning approach, we conduct experiments on a set of graph and non-graph datasets (explained in Sec. \ref{subsec:datasets}). The implementation details are presented in  Sec. \ref{subsec:implementation}. We compare the results achieved by GRATIS with existing state-of-the-art methods on all employed datasets in Sec. \ref{subsec:sota}. We also systematically conduct a series of ablation studies to individually evaluate the effectiveness of each component in the GRATIS in Sec. \ref{subsec:ablation}.

\subsection{Datasets}
\label{subsec:datasets}

This paper evaluates the proposed approach on three typical graph analysis tasks: graph classification, vertex (node) classification and edge (link) prediction, where two datasets are employed for each task.

\noindent \textbf{Graph datasets:} Six graph datasets are employed (two for each task): (i) \textbf{Graph classification:} the MNIST \cite{lecun1998gradient} and CIFAR10 \cite{krizhevsky2009learning} datasets are employed; (ii) \textbf{Vertex (node) classification:} the PATTERN \cite{dwivedi2020benchmarking} and CLUSTER \cite{dwivedi2020benchmarking} datasets are employed; and (iii) \textbf{Edge link prediction:} the TSP \cite{dwivedi2020benchmarking} and COLLAB \cite{hu2020open} datasets are employed. All datasets are provided in \cite{dwivedi2020benchmarking}, and we follow the same protocol as \cite{dwivedi2020benchmarking} to pre-process and organize each of these datasets for its corresponding graph analysis task.

\noindent \textbf{Non-graph datasets:} Five non-graph face image datasets are also employed for three types of graph analysis tasks: (i) \textbf{graph classification/regression:} the FER 2013 \cite{sambare_2020} and RAF-DB \cite{li2017reliable} facial expression recognition (FER) datasets are employed, where we produce a graph for each face image, and predict image-level facial expressions (i.e., seven class classification problem) based on the produced graph. In addition, the AVEC 2019 depression dataset \cite{ringeval2019avec} is also employed, where each subject's spatio-temporal behaviour is represented by a multi-channel time-series data. The target is to predict a depression severity value (PHQ-8 score) for each subject; (ii) \textbf{Vertex (node) classification:} the BP4D \cite{zhang2014bp4d} and DISFA \cite{mavadati2013disfa} Facial Action Units (AUs) Recognition datasets are employed, where the task is to jointly predict multiple AUs' activation from each face image (i.e., multi-task binary classification problem). Specifically, we produce a graph for each face image, each of whose vertex describes a specific AU's activation in the target face image; and (iii) \textbf{Link prediction:} the BP4D and DISFA datasets are again employed, where we aim to recognize the co-occurrence pattern between a pair of AUs (vertices), i.e., the edge pattern of the corresponding AUs. In this paper, we define the AUs' co-occurrence recognition (AU-based link prediction) as a four-class classification problem, \emph{i.e.,} for a pair of nodes $\mathbf{v}_i$ and $\mathbf{v}_j$: (1) both $\mathbf{v}_i$ and $\mathbf{v}_j$ are inactivated; (2) $\mathbf{v}_i$ is inactivated and $\mathbf{v}_j$ is active; (3) $\mathbf{v}_i$ is active and $\mathbf{v}_j$ is inactivated; or (4) both $\mathbf{v}_i$ and $\mathbf{v}_j$ are active.

\subsection{Implementation details}
\label{subsec:implementation}

\noindent In this paper, we evaluated the GRATIS based on two widely-used GNN models that can process multi-dimensional edge features: GatedGCN \cite{bresson2017residual} and GAT \cite{velivckovic2017graph}. We also individually evaluate two standard deep learning models: ResNet \cite{he2016deep} and Swin-Transformer \cite{liu2021Swin}, as the backbone for non-graph dataset-based experiments.

\subsubsection{Experiments on Graph data}

\noindent For all graph dataset-based experiments, we follow the same data splits and training settings in \cite{dwivedi2020benchmarking}. The employed GAT and GatedGCN have 4 GNN layers for experiments on MNIST and CIFAR10 datasets, 16 GNN layers for experiments on PATTERN, CLUSTER and TSP datasets, and 3 GNN layers for experiments on the COLLAB dataset. For all experiments on graph data, we use Adam optimizer \cite{kingma2014adam} with a learning rate decay strategy to train all our models. To make fair comparison and to enable the experiments to be reproducible, there is no extra pre-processing, or post-processing steps for graph data-based experiments. The implementation details of compared baselines (e.g., GCN, GAT, and GatedGCN) can be found in \cite{dwivedi2020benchmarking}. The detailed hyper-parameter settings for all graph and non-graph dataset-based experiments are provided in the supplementary material.

\subsubsection{Experiments on non-graph data}

\textbf{Pre-processing:} For all non-graph facial AU and expression analysis experiments, we only perform a standard and generic pre-processing that has been used by almost all previous works, which applies MTCNN \cite{yin2017multi} to detect and align face for each frame, and then crop/re-shape it to $224 \times 224$ as the input for backbones.

\noindent \textbf{Model settings and training details:} For both FER and AU recognition experiments, the rule $\mathcal{R}$ in Eqa. \ref{eq:raw_a_nongraph} is set such that each vertex connects to all vertices in the basic graph. For all \textbf{FER experiments}, we follow the same training and testing splits defined in FER 2013 and RAF-DB datasets, where four vertices are used to construct facial graphs. Meanwhile, we follow \cite{xu2021two} to construct the initial spectral graph of multi-channel time-series facial behaviour data in GD module for each clip on AVEC 2019 depression dataset, and then applies our TTP and MEFG to further optimize vertex feature, topology and edge features to construct final graph for \textbf{depression recognition}. We then follow the same protocol as previous studies \cite{zhao2016deep,li2018eac,Song_2021_CVPR} to conduct subject-independent 3-folds cross-validation on each \textbf{AU recognition} dataset, and report the averaged results over 3 folds. The number $C$ for choosing nearest neighbors in the FGG is set to 3 and 4 for BP4D and DISFA, respectively.
For AU co-occurrence experiments, we also follow the same cross-validation splits from the AU recognition experiments by conducting subject-independent 3-fold cross-validation for both BP4D and DISFA datasets. We additionally employ a three-layer multi-layer perceptron (MLP) as the final classifier on the top of the final GNN layer to predict the co-occurrence of each AU pair. The results reported for both AU recognition and AU co-occurrence detection are achieved by averaging the results of 3-fold cross-validation. The initial weights of backbone models are obtained from ImageNet \cite{deng2009imagenet}, while the initial weights of AU co-occurrence experiments are obtained from corresponding AU recognition datasets (i.e., either BP4D or DISFA). For all experiments, we employ a vanilla GCN \cite{henaff2015deep} for the TTP module. The AdamW optimizer \cite{loshchilov2017decoupled} is employed for training all models and the cosine decay learning rate scheduler is also used. All our experiments were conducted using the open-source PyTorch platform.

\vspace{0.02cm}
\subsection{Evaluation metrics}

\textbf{Graph data-based experiments:} To evaluate the performances of approaches on six graph datasets, we follow the same protocol as explained in \cite{dwivedi2020benchmarking}. In particular, the graph classification results on MNIST and CIFAR10 are represented using the classification accuracy, and the vertex (node) classification results on PARTTERN and CLUSTER are achieved by averaging the classification accuracy of all vertices. Finally, we employ Hits$@$50 \cite{hu2020open} to evaluate the link prediction result on COLLAB, while the link prediction result on TSP is evaluated using the $F1$ score (i.e., $F1 = 2 \frac{P \cdot R}{P+R}$), which takes both the recognition precision $P$ and recall rate $R$ into consideration.

\textbf{Non-graph data-based experiments:} We use the classification accuracy as the evaluation metric for FER tasks (graph classification) conducted on FER 2013 and RAF-DB, while Concordance Correlation Coefficient (CCC) is employed as the metrics for depression recognition. The AU recognition (vertex classification) performances are measured by averaging frame-based $F1$ score as it has been frequently used in previous AU recognition studies \cite{song2021uncertain,shao2021jaa,jacob2021facial,yang2021exploiting}. Finally, since the label distributions of AU co-occurrence patterns on both datasets are quite unbalanced, the AU co-occurrence (link prediction) experiments are evaluated using Unweighted Average Recall (UAR) which is defined as the unweighted average of the class-specific recalls on all classes.



\subsection{Comparison to state-of-the-art approaches}
\label{subsec:sota}

\subsubsection{Experimental results on graph datasets}

\noindent We first compare the graph analysis results achieved by our best systems with recent graph analysis methods in Table \ref{tb:SOTA_graph}, where two GNNs are individually employed as the predictor to process graphs produced by our approach. It can be seen that graphs enhanced by our approach (i.e., graphs with deep-learned multi-dimensional edge features) achieved the state-of-the-art results on all six graph datasets when GatedGCN is employed as the classifier, where the improvements on two link prediction datasets are consistently over $2\%$, indicating that the multi-dimensional edge features deep-learned by our approach provide better description of the task-specific relationship between vertices. More importantly, it is clear that when using GatedGCN and GAT as the predictor, the graphs enhanced by our approach show clear advantages over the original graphs on all datasets, i.e., the graphs enhanced by the proposed approach provided $6.5\%$ (GAT) and $3.5\%$ (GatedGCN) average relative improvements in comparison to original graphs. Since the GatedGCN baseline processes multi-dimensional edge features while GAT baseline made predictions based on single-value edge graphs, the improvements of our approach on GAT is much higher. These results provide a solid evidence that our approach can further exploit the underlying task-specific cues from pre-defined graphs, and produce strong/rich edge representations to enhance these pre-defined graphs for various graph analysis tasks, regardless of the employed GNN predictor.

\begin{table*}[t]
\caption{Comparison to the state-of-the-art systems on six medium-scale graph benchmark datasets provided by \cite{dwivedi2020benchmarking}, where $\star$ indicates the results reported from \cite{dwivedi2020benchmarking}, and the numbers in brackets indicate the relative improvements of the graphs enhanced by our approach over the original graphs when GAT and GatedGCN are individually used as the predictor.  The best, second best, and third best results of each column are indicated with brackets and bold font, brackets alone, and underline, respectively.}
\label{tb:SOTA_graph}
\centering
\resizebox{0.8\linewidth}{!}{
\begin{tabular}{lcccccc}
\toprule
\textbf{Task} &\multicolumn{2}{c}{\textbf{Graph classification}} &\multicolumn{2}{c}{\textbf{Vertex classification}} &\multicolumn{2}{c}{\textbf{Link prediction}} \\
\toprule
{\textbf{Dataset}} &{\textbf{MNIST}} &{\textbf{CIFAR10}} &\textbf{PATTERN} &\textbf{CLUSTER} &\textbf{TSP} &\textbf{COLLAB} \\
\toprule
\textbf{Method} &\multicolumn{2}{c}{\textbf{Test Acc($\%$)}$\uparrow$} &\multicolumn{2}{c}{\textbf{Test Acc($\%$)}$\uparrow$} &\textbf{Test $F_1$} $\uparrow$ &\textbf{Test Hits} $\uparrow$ \\
\toprule
{GCN$^{\star}$ \cite{kipf2016semi,dwivedi2020benchmarking}} &{90.71} &{56.34} &{71.89} &{68.50} &{0.631} &{50.42} \\
{GIN$^{\star}$ \cite{xu2018powerful,dwivedi2020benchmarking}} &{96.49} &{55.26} &{85.59} &{64.72} &{0.660} &{41.73}\\
{GAT$^{\star}$ \cite{velivckovic2017graph,dwivedi2020benchmarking}} &{95.54} &{64.22} &{78.27} &{70.59} &{ 0.671} &{51.50}\\
{GatedGCN$^{\star}$ \cite{bresson2017residual,dwivedi2020benchmarking}} &{97.34} &{67.31} &{86.51} &76.08 &\underline{0.808} &\underline{52.64}\\
{PNA \cite{corso2020principal}} &{97.69} &\underline{70.35} &{86.57} &{-} &{-} &{-}\\
{EGT \cite{hussain2021edge}} &\underline{97.72} &{67.00} &86.83 &\textbf{77.91} &\textbf{0.810} &{-}\\
{DGN \cite{beaini2021directional}} &{-} &\textbf{72.84} &{86.68} &{-} &{-} &{-}\\
{GNAS-MP \cite{cai2021rethinking}} &\textbf{98.01} &{70.10} &\textbf{86.85} &{74.77} &{0.742} &{-}\\
SAT \cite{chen2022structure} &{-} &{-} &\underline{86.84} &\underline{77.86} &{-} &{-} \\
\toprule
{Ours(GAT)}  &{96.97 ($+1.5\%$)} &{66.36 ($+3.3\%$)} &{81.04 ($+3.5\%$)} &{74.84 ($+6.0\%$)} &{0.807 ($+20.3\%$)} &\textbf{53.68} ($+4.2\%$) \\ 
{Ours (GatedGCN)} &[\textbf{98.18}]($+0.9\%$) &[\textbf{73.84}]  ($+9.7\%$) &[\textbf{86.86]}]  ($+0.4\%$) &[\textbf{78.65}] ($+3.4\%$) &[\textbf{0.830}]  ($+2.7\%$) &[\textbf{54.76}]  ($+4.0\%$)\\
\toprule
\end{tabular}
}
\end{table*}

\subsubsection{Experimental results on non-graph datasets}

\noindent We then compare the results achieved by our best systems with several state-of-the-art methods on non-graph tasks. According to Table \ref{tb:fer_soat}, when applying our approach to build graphs from the latent representation produced by the standard CNN (ResNet) and Transformer (Swin-Transformer), the FER (graph classification) performances on both FER2013 and RAF-DB datasets have been clearly improved compared to the original ResNet and Swin-Transformer, where both systems achieved superior results to other existing methods on FER 2013 and comparable performances to the state-of-the-art on RAF-DB. Meanwhile, our approach also achieved the state-of-the-art performances for FAU feature and ResNet feature-based systems (each feature is represented as a multi-channel time-series) on AVEC 2019 depression dataset, with $5.6\%$ and $6.9\%$ average improvements, respectively. More recently, an approach have employed our MEFG to non-graph data-based audio analysis (non-graph audio data-based graph classification) \cite{hou2022multi}. The results show that MEFG largely enhanced the corresponding baseline systems (i.e., the accuracy is improved from $59.7\%$ to $78.1\%$), and achieved the state-of-the-art performance (i.e., $78.1\%$ compared to $76.6\%$ achieved by the previous SOTA).

Table~\ref{ex:tab_BP4D_sota} and Table~\ref{ex:tab_DISFA_sota} report the AU occurrence recognition results. For fair comparisons, we only compare our approach with static face-based methods that did not remove any frame from the datasets. The results show that the proposed graph encoding approach allows both backbones (ResNet-50 and Swin Transformer-Base (Swin-B)) to achieve superior overall $F1$ scores to all other listed approaches, with $0.5\%$ and $1.3\%$ average improvements over the state-of-the-art \cite{jacob2021facial} on the BP4D dataset and $1.6\%$ and $2.4\%$ average improvements on the DISFA dataset. Specifically, our approach allows both backbones to achieve the top three performances for $9$ out of $12$ AUs' recognition (\emph{e.g.}, AU 4, AU 6, AU 7, AU 10, AU 12, AU 14, AU 15, AU 17, and AU 23) among all listed approaches on the BP4D dataset, and each backbone to achieve the top three performances for $5$ out of $8$ AUs' recognition on DISFA dataset.

The aforementioned results suggest that the proposed graph representation learning approach can exploit more task-related cues from latent features of the non-graph data extracted from backbones, and construct task-specific graph representations for various graph analysis tasks, which can further improve the non-graph data analysis performances. Since our models were built without any complex pre-processing or specific training tricks, these results indicate that it has a good generalization capability on different non-graph facial datasets. In other words, our approach has a potential to push up the upper-bound of different facial analysis performance if more domain-specific latent features (backbones) or loss functions are employed. We additionally provide the AUC results achieved for AU recognition as well as the confusion matrices achieved for facial expression recognition and AU co-occurrence recognition in the supplementary material.

\begin{table*}[t]
\setlength{\extrarowheight}{1pt}
\caption{Comparison between our best systems and other approaches on non-graph data-based graph classification tasks, where both facial expression recognition (on FER2013 and RAF-DB datasets), and depression recognition (on AVEC 2019 dataset) were conducted. Specifically, each sample of the FER task is a static face image, while each sample of the AVEC 2019 dataset is represented by a multi-channel time-series signal (FAU feature or ResNet feature), where each channel is treated as a vertex, and the global contextual representation is defined as the matrix that concatenates all channels of time-series signals.}
\label{tb:SOTA_nongraph_classification}
\centering
\small
\resizebox{0.66\linewidth}{!}{
\begin{tabular}{lccccc}
\cline{1-3} \cline{5-6}
\multicolumn{3}{c}{\textbf{Facial Expression Recognition}} &&\multicolumn{2}{c}{\textbf{Depression Recognition}} \\
\cline{1-3} \cline{5-6}
&{\textbf{FER2013}} &{\textbf{RAF-DB}} &  & &  \textbf{AVEC 2019} \\
\cline{1-3} \cline{5-6}
\textbf{Method} &\multicolumn{2}{c}{\textbf{Test Acc($\%$)}$\uparrow$} & &  \textbf{Method} & \textbf{CCC} $\uparrow$\\
\cline{1-3} \cline{5-6}
Hasani et al. \cite{hasani2020breg} & 71.53  & - &  & Ringeval et al.(FAU) \cite{ringeval2019avec} & 0.019 \\
Shi et al. \cite{shi2021facial} & 71.54  & -  &  & Song et al.(FAU) \cite{song2020spectral} & 0.330 \\
Shao et al. \cite{shao2019three} & 71.14 & - &  & Xu et al.(FAU)  \cite{xu2021two} & 0.380 \\
\cline{5-6}
Vulpe et al. \cite{vulpe2021convolutional} & 72.16 & - &  & Ours (TTP-Graph-FAUs) & \underline{0.391} \\
Florea et al. \cite{florea2019annealed} & - & 84.50 &  & Ours (MEFG-Graph-FAUs)  & \textbf{0.415} \\
Li et al. \cite{li2018occlusion}  & - &  85.07 &  & Ours (MEFG-TTP-Graph-FAUs) & [\textbf{0.436}] \\
\cline{5-6}
Zhang et al. \cite{zhang2021learning} & - & 86.40 &  &   Ringeval et al.(ResNet) \cite{ringeval2019avec} &  0.120  \\
Wang et al. \cite{wang2020suppressing} & - & \underline{87.03} &  & Song et al.(ResNet) \cite{song2020spectral} & 0.169 \\
Mahmoudi et al. \cite{mahmoudi2022kernel} & 71.08 & [\textbf{87.84}] &  & Xu et al.(ResNet) \cite{xu2021two} & 0.186 \\ 
\cline{5-6}
Fard et al. \cite{fard2022ad} & 72.03 & 86.96 &  & Ours (TTP-Graph-ResNet) & \underline{0.213}\\  
\cline{1-3}
Ours (ResNet) & [73.15] & [87.13] &  & Ours (MEFG-Graph-ResNet) & \textbf{0.236} \\
Ours (Swin-B) & [\textbf{73.56}] & 86.70 &  & Ours (MEFG-TTP-Graph-ResNet)  & [\textbf{0.255}]\\
\cline{1-3} \cline{5-6}
\end{tabular}
}
\label{tb:fer_soat}
\end{table*}

\textbf{Summary:} The achieved results provide solid evidences that the proposed graph representation learning approach can clearly improve performances for various graph and non-graph data analysis tasks. Particularly, our approach is robust to different backbone feature extractors, i.e., it enhances the performances of both backbones. Subsequently, we conclude that the proposed approach can produce a graph representation that contains a task-specific topology and multi-dimensional edge features for various pre-defined graph and non-graph data.


\begin{table*}[t]
    \caption{$F1$ scores (in \%) achieved for 12 AUs on BP4D dataset, where the three methods (SRERL, UGN-B and HMP-PS) listed in the middle of the table are also built with graphs. The best, second best, and third best results of each column are indicated with brackets and bold font, brackets alone, and underline, respectively.
    }
    \centering
     \small
    \resizebox{0.86\linewidth}{!}{        \begin{tabular}{lccccccccccccc}
        
        \Xhline{3\arrayrulewidth}
        \multicolumn{1}{l}{\multirow{2}{*}{Method}} & \multicolumn{12}{c}{AU}  & \multirow{2}{*}{\textbf{Avg.}} \\ \cmidrule(lr){2-13}
        \multicolumn{1}{l}{}    & 1    & 2    & 4    & 6   & 7   & 10   & 12   & 14   & 15   & 17   & 23   & 24   &      \\ \midrule
        DRML  \cite{zhao2016deep}                  &36.4  &41.8  &43.0  &55.0   &67.0   &66.3  &65.8   &54.1   &33.2  &48.0  &31.7  &30.0  &48.3 \\
        EAC-Net \cite{li2018eac}                &39.0  &35.2  &48.6  &76.1   &72.9   &81.9  &86.2   &58.8   &37.5  &59.1  &35.9  &35.8  &55.9  \\
        JAA-Net  \cite{shao2018deep}               &47.2  &44.0  &54.9  &77.5   &74.6   &84.0  &86.9   &61.9   &43.6  &60.3  &42.7  &41.9  &60.0  \\

    
        LP-Net  \cite{niu2019local}                &43.4  &38.0  &54.2  &77.1   &76.7   &83.8  &87.2   &63.3   &45.3  &60.5  &48.1  &54.2  &61.0  \\
        ARL   \cite{shao2019facial}                  &45.8  &39.8  &55.1  &75.7   &77.2   &82.3  &86.6   &58.8   &47.6  &62.1  &47.4  &[55.4]  &61.1  \\
        SEV-Net \cite{yang2021exploiting}                &[\textbf{58.2}]  &[\textbf{50.4}]  &58.3  &[\textbf{81.9}]  &73.9   &[\textbf{87.8}]  &87.5   &61.6   &[52.6]  &62.2  &44.6  &47.6  &63.9  \\
        FAUDT \cite{jacob2021facial}                   &51.7  &[49.3]  &[\textbf{61.0}]  &77.8   &\underline{79.5}   &82.9  &86.3   &[67.6]   &51.9  &63.0  &43.7  &[\textbf{56.3}]  &\underline{64.2}   \\\midrule
    
        SRERL \cite{li2019semantic}                  &46.9  &45.3  &55.6  &77.1   &78.4   &83.5  &\underline{87.6}   &63.9   &52.2  &[63.9]  &47.1  &53.3  &62.9  \\
        UGN-B \cite{song2021uncertain}                  &[54.2]  &46.4  &56.8  &76.2   &76.7   &82.4  &86.1   &64.7   &51.2  &63.1  &48.5  &53.6  &63.3  \\
        HMP-PS \cite{Song_2021_CVPR}                 &53.1  &46.1  &56.0  &76.5   &76.9   &82.1  &86.4   &64.8   &51.5  &63.0  &[49.9]  &54.5  &63.4  \\ \midrule

        Ours (ResNet)           &\underline{53.7}      &\underline{46.9}      &\underline{59.0}      &\underline{78.5}       &[80.0]       &\underline{84.4}      &[87.8]       &\underline{67.3}       &\underline{52.5}   &\underline{63.2}   &\textbf{50.6}  &52.4  &[64.7]     \\\
        Ours (SwinB)              &52.7  &44.3  &[60.9]  & [79.9]   &[\textbf{80.1}]   &[85.3]  &[\textbf{89.2}]   &[\textbf{69.4}]   &[\textbf{55.4}]  &[\textbf{64.4}]  &\underline{49.8}  &\underline{55.1}  &[\textbf{65.5}]  \\
        \Xhline{3\arrayrulewidth}
        \end{tabular}}
    \label{ex:tab_BP4D_sota}
\end{table*}

\begin{table}[t]
    \caption{$F1$ scores (in \%) achieved for 8 AUs on DISFA dataset. The best, second best, and third best results of each column are indicated with brackets and bold font, brackets alone, and underline, respectively.}
    \centering
    \small
    \resizebox{\linewidth}{!}{
    \begin{tabular}{lccccccccc}
    \Xhline{3\arrayrulewidth}
    \multicolumn{1}{l}{\multirow{2}{*}{Method}} & \multicolumn{8}{c}{AU}  & \multirow{2}{*}{\textbf{Avg.}} \\ \cmidrule(lr){2-9}
    \multicolumn{1}{l}{}    & 1               & 2       & 4   & 6   & 9   & 12   & 25   & 26     &      \\\midrule
    DRML \cite{zhao2016deep}                    &17.3             &17.7     &37.4  &29.0   &10.7   &37.7  &38.5   &20.1   &26.7 \\
    EAC-Net \cite{li2018eac}                 &41.5             &26.4     &66.4  &50.7   &[\textbf{80.5}]   &[\textbf{89.3}]  &88.9   &15.6   &48.5   \\
    JAA-Net \cite{shao2018deep}                 &43.7             &46.2     &56.0  &41.4   &44.7   &69.6  &88.3   &58.4   &56.0   \\

    LP-Net \cite{niu2019local}                 &29.9              &24.7              &72.7  &46.8   &49.6   &72.9  &93.8   &65.0   &56.9   \\
    ARL \cite{shao2019facial}                    &43.9              &42.1              &63.6  &41.8   &40.0   &76.2  &[95.2]   &[66.8]   &58.7   \\
    SEV-Net  \cite{yang2021exploiting}                &\textbf{[55.3]}     &[\textbf{53.1}]     &61.5  &\underline{53.6}   &38.2   &71.6  &[\textbf{95.7}]   &41.5   &58.8   \\
    FAUDT \cite{jacob2021facial}                  &46.1              &\underline{48.6}  &\underline{72.8}  &[\textbf{56.7}]   &50.0   &72.1  &90.8   &55.4   &\underline{61.5}   \\\midrule
    
    SRERL \cite{li2019semantic}                  &45.7              &47.8  &59.6  &47.1   &45.6   &73.5  &84.3   &43.6   &55.9   \\
    UGN-B \cite{song2021uncertain}                  &43.3              &48.1  &63.4  &49.5   &48.2   &72.9  &90.8   &59.0   &60.0   \\
    HMP-PS  \cite{Song_2021_CVPR}                 &38.0              &45.9  &65.2  &50.9   &\underline{50.8}   &76.0  &93.3   &[\textbf{67.6}]   &61.0   \\ \midrule

    Ours (ResNet)              &[54.6]  &47.1  &[72.9]  &[54.0]   &[55.7]   &\underline{76.7}  &91.1   &53.0  &[63.1]     \\
    Ours (SwinB)              &47.9     &[48.7]      &[\textbf{75.0}]      &51.2       &51.6       &[77.4]      &\underline{94.2}      &\underline{65.2}      &[\textbf{63.9}]    \\
    \Xhline{3\arrayrulewidth}
    \end{tabular}
    }
    \label{ex:tab_DISFA_sota}
\end{table}

\subsection{Ablation studies}
\label{subsec:ablation}

\noindent In this section, we specifically: (i) evaluate the benefits brought by the proposed TTP and MEFG modules on different tasks; and (ii) evaluate the robustness of our approach to different parameter settings when conducting different graph analysis tasks.

\subsubsection{Benefits of the TTP and MEFG modules}

\noindent Table \ref{tb:Ablation_graph}, Table \ref{tb:Ablation_nongraph_Res} and Table \ref{tb:Ablation_nongraph_Swin} report the results achieved by different edge settings using two different GNN predictors. Specifically, for graph datasets, we evaluated graphs that have: (i) the original graph topology and edge features manually defined in \cite{dwivedi2020benchmarking}; (ii) and (iii) the graph topology that is defined by connecting each vertex to its $C$ nearest vertices using either L1 or Mahalanobis as the measurement, where L1 or Mahalanobis distance between vertices is also employed as the single-value edge feature; (iv) the task-specific graph topology decided by TTP, and edge features defined using the same methods of \cite{dwivedi2020benchmarking}; (v) the original graph topology that is manually defined in \cite{dwivedi2020benchmarking}, and multi-dimensional edge features learned by MEFG; and (vi) the task-specific graph topology decided by TTP, and multi-dimensional edge features learned by MEFG.

We generate several graph representations for each non-graph sample from the latent representation produced by the backbone based on: (ii) GD and TTP modules, i.e., graphs that have task-specific topology (For non-graph samples, their initial graphs are jointly defined by GD and TTP modules); (iii) and (iv) GD and TTP modules as well as hand-crafted single-value edge features (i.e., L1 and L2 distances between vertices); (v)/(vii) GD and MEFG modules, i.e., graphs that have task-specific multi-dimensional edge features, where each vertex is connected to all other vertices; and (vi)/(ix) GD, TTP and MEFG modules, i.e., graphs that have task-specific topology and multi-dimensional edge features. We compare these graphs with the results achieved fro corresponding backbones (i.e., the setting (i) in both Table \ref{tb:Ablation_nongraph_Res} and Table \ref{tb:Ablation_nongraph_Swin}, where no graph learning module is used). Here, we do not specifically report the results achieved for backbone-GD for non-graph data-based experiments as the GD module only simply treats the latent features produced from the backbone as vertex features for the graph, and fully connects them without conducting any feature learning. Thus, the results achieved by backbone-GD is almost the same to the results achieved by the backbone.

\begin{figure}[t]
  \centering
  \includegraphics[width=1.0\columnwidth]{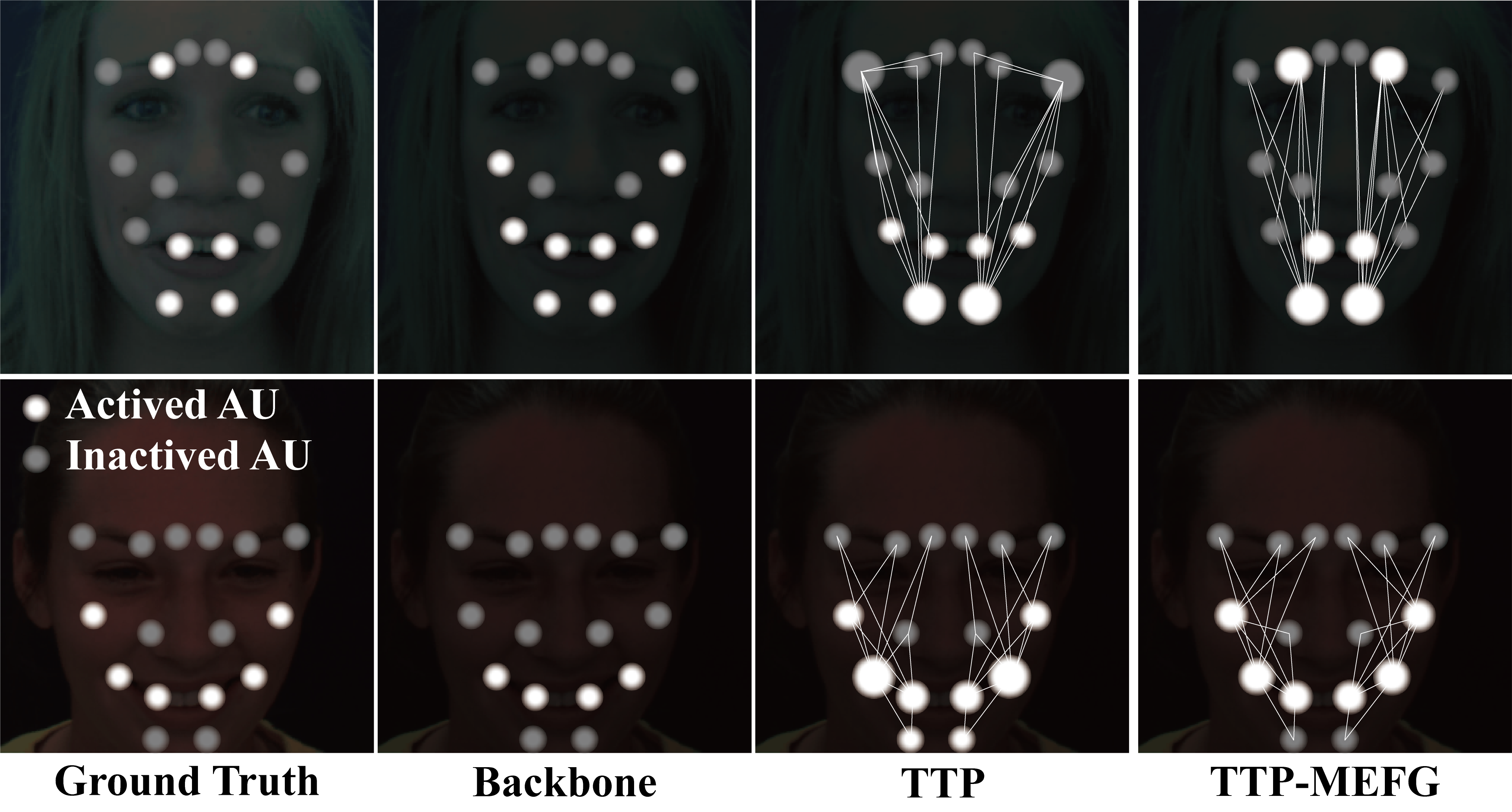}
  \caption{Visualization of association cues encoded in vertex features (only systems of the last two columns encode such cues). We connect each vertex to its $C$ nearest neighbours, where vertices of activated AUs usually have more connections than vertices of inactivated AUs. Systems using such relationship cues have enhanced AU recognition results (predictions of the column 3 is better than the column 2).}
\label{fig:Visualization_AU}
\end{figure}

\textbf{Benefits of the TTP:} According to Table \ref{tb:Ablation_graph}, we found that graphs whose typologies are enhanced by the TTP achieved superior results than graphs whose typologies are manually defined using either prior knowledge \cite{dwivedi2020benchmarking}, or other pre-defined rules (e.g., L1 and Mahalanobis distances) on all evaluated datasets, regardless of the employed GNN predictor. Meanwhile, it can be observed from Table. \ref{tb:Ablation_nongraph_Res} and Table. \ref{tb:Ablation_nongraph_Swin} that the graph representations produced by our TTP module also achieved clearly better performances on all non-graph datasets in comparison to latent features produced from backbones. In addition, Table. \ref{tb:fer_soat} also illustrates that based on the same vertex features, the graph topology learned by TTP allow the produced graph achieving clear improvements in depression recognition over the fully connected graph \cite{xu2021two}. These results not only demonstrate that the proposed TTP module can assign better typologies that contain more task-specific cues for pre-defined graphs, but also show that it can build strong graph representations from non-graph data, which is a better way than the latent representations learned from the backbone, for describing task-specific cues of the non-graph data. Fig. \ref{fig:Visualization_AU} visualized the influence of the TTP on building facial graph representations for AU recognition. When a facial AU is active, its movement usually affects other facial regions (\emph{i.e.}, the activation of other AUs) while inactivated AUs would not have such an effect. As a result, our TTP module simulates this phenomenon by connecting active AUs to all other AUs (including active and inactivated AUs).

\begin{table}[t]
\caption{Results achieved for different TTP and MEFG settings on graph datasets, where GatedGCN or GAT is employed as the classifier. Since graphs in PATTERN and CLUSTER consist of single-value vertex features, the M-dist denotes the Mahalanobis distance, which is not applicable to measure the distance between vertices whose dimensions are one.}
\label{tb:Ablation_graph}
\centering
\resizebox{\linewidth}{!}{
\begin{tabular}{lcccccc}
\toprule
\multirow{2}*{\textbf{Method}} &\multicolumn{2}{c}{\textbf{Graph classification}} &\multicolumn{2}{c}{\textbf{Vertex classification}} &\multicolumn{2}{c}{\textbf{Link prediction}}\\
\cline{2-7} 
 & \textbf{MNIST} & \textbf{CIFAR10} &\textbf{PATTERN} & \textbf{CLUSTER} & \textbf{TSP} & \textbf{COLLAB} \\
\toprule
\multicolumn{7}{l}{\textsl{(A) Predictor: GatedGCN} } \\ 
(i) Original \cite{dwivedi2020benchmarking} &97.34 &67.31 &86.51 &76.08 &0.808 &52.64  \\
(ii) L1 &{96.92} &{68.19} &{86.04} &{75.51} &{-} &{-} \\ 
(iii) M-dist &{94.24} &{68.61} &{-} &{-} &{-} &{-} \\ 
(iv) TTP &97.49 &68.38 &86.61 &78.43 &{-} &{-} \\
(v) MEFG &\textbf{98.18} &\textbf{73.84} &86.72 &78.30 &\textbf{0.830} &\textbf{54.76} \\ 
(vi) TTP-MEFG &{97.96} &{72.53} &\textbf{86.86} &\textbf{78.65} &{-} &{-} \\
\toprule
\multicolumn{7}{l}{\textsl{(B) Predictor: GAT} } \\ 
(i) Original \cite{dwivedi2020benchmarking} &95.54 &64.22 &78.27 &70.59 &0.671&51.50 \\ 
(ii) L1 &96.57 &65.23 &{78.31} &{70.91} &{-} &{-} \\ 
(iii) M-dist &{95.27} &{63.89} &{-} &{-} &{-} &{-} \\ 
(iv) TTP &{96.15} &{65.84} &{78.53} &{74.23} &{-} &{-} \\
(v) MEFG &96.88 &66.07 &\textbf{81.04} &74.89 &\textbf{0.807} &\textbf{53.68} \\
(vi) TTP-MEFG &\textbf{96.97} &\textbf{66.36} &80.93 &\textbf{74.84} &{-} &{-} \\ 
\toprule
\end{tabular}
}
\end{table}

\begin{table}[t]
\caption{Results achieved for different TTP and MEFG settings on non-graph face image datasets, where GatedGCN or GAT are employed as the classifier and ResNe \cite{he2016deep} is employed as the backbone.}
\label{tb:Ablation_nongraph_Res}
\centering
\resizebox{\linewidth}{!}{
\begin{tabular}{lcccccc}
\toprule
\multirow{2}*{\textbf{Method}} &\multicolumn{2}{c}{\textbf{Graph classification}} 
&\multicolumn{2}{c}{\textbf{Vertex classification}} &\multicolumn{2}{c}{\textbf{Link prediction}}\\
\cline{2-7} 
&{\textbf{FER2013}} &{\textbf{RAF-DB}} &\textbf{BP4D} &\textbf{DISFA} &\textbf{BP4D} &\textbf{DISFA} \\
\toprule
(i) ResNet \cite{he2016deep}  & 72.22 & 85.98&61.8 &58.2 & 57.8    & 67.1\\
(ii) B-GD-TTP  & 73.06  &86.79 &63.7 &61.3 &{-} &{-} \\
(iii) B-GD-TTP-L1 & 71.19& 83.96&63.7  &\textbf{63.7} &{-} &{-} \\
(iv) B-GD-TTP-L2 &72.05 &84.80 &63.1 &\textbf{63.7} &{-} &{-} \\
(v) B-GD-MEFG (Gated)  & 72.78 &86.57 &63.9 &61.1 &\textbf{60.6} & \textbf{69.3}\\
(vi) B-GD-TTP-MEFG (Gated)  & \textbf{73.15} &87.06 &\textbf{64.7} &63.1 &{-} &{-} \\
(vii) B-GD-MEFG (GAT)  &72.64 &86.51 &63.7 &62.4 & 60.2 & 68.4 \\
(viii) B-GD-TTP-MEFG (GAT)  &73.00 & \textbf{87.13} &64.2 &63.5 &{-} &{-} \\
\hline
\end{tabular}
}
\end{table}


\begin{table}[t]
\caption{Results achieved for different TTP and MEFG settings on non-graph face image datasets, where GatedGCN or GAT are employed as the classifier and Swin-Transformer \cite{liu2021Swin} is employed as the backbone.}
\label{tb:Ablation_nongraph_Swin}
\centering
\resizebox{\linewidth}{!}{
\begin{tabular}{lcccccc}
\toprule
\multirow{2}*{\textbf{Method}} &\multicolumn{2}{c}{\textbf{Graph classification}} 
&\multicolumn{2}{c}{\textbf{Vertex classification}} &\multicolumn{2}{c}{\textbf{Link prediction}}\\
\cline{2-7} 
{} &{\textbf{FER2013}} &{\textbf{RAF-DB}} &\textbf{BP4D} &\textbf{DISFA} &\textbf{BP4D} &\textbf{DISFA} \\
\toprule
(i) Swin-Transformer \cite{liu2021Swin}  &70.06 & 84.91&63.9 &58.7 & 58.2 & 65.1 \\
(ii) B-GD-TTP  & 73.35&85.98 &65.1 &62.1 &{-} &{-} \\
(iii) B-GD-TTP-L1 &72.05 & 83.60 &64.5 &63.6 &{-} &{-} \\
(iv) B-GD-TTP-L2 &72.49 &84.27 &64.4 &63.6 &{-} &{-} \\
(vi) B-GD-MEFG (Gated) &72.89 & 85.51&64.6 &62.1 & 61.1 & 71.7\\
(vii) B-GD-TTP-MEFG (Gated) &\textbf{73.56} & \textbf{86.70} &\textbf{65.5} &\textbf{63.9} &{-} &{-} \\
(viii) B-GD-MEFG (GAT)  & 72.69&85.37 &64.8 &62.6 &\textbf{61.8} & 72.6 \\
(ix) B-GD-TTP-MEFG (GAT)  &73.22 & 86.44 &65.4 &63.6 &{-} &{-} \\
\toprule
\end{tabular}
}
\end{table}

\textbf{Benefits of the MEFG:} It can be observed that simply applying the proposed MEFG module to assign a pair of multi-dimensional edge features for each manually-defined edge leads to large performance gains for all systems on all graph and almost all non-graph tasks (except ResNet-GD-TTP-L1 and ResNet-GD-TTP-L2 systems achieved slightly better performance on DISFA dataset), compared to their baselines and systems used other single-value edges (i.e., the settings (ii) and (iii) used for graph datasets and settings (iii) and (iv) used for graph datasets), where the result of the GAT-based system on the TSP graph dataset is boosted from $0.671$ to $0.807$ and the AU recognition $F1$ result of the GAT-based system on the DISFA non-graph dataset is boosted from $58.7$ to $62.6$. Importantly, our approach achieved clear improvements for different backbones and GNN predictors. As shown in Table. \ref{tb:fer_soat}, the MEFG also largely enhanced the fully connected graph \cite{xu2021two} in depression recognition.  These results indicate that the relationship between a pair of vertices whose representation are multi-dimensional vectors could be better described by deep-learned multi-dimensional edge features. More specifically, our deep-learned multi-dimensional edge features can provide more informative cues than single-value edge features (i.e., the number of values to describe relationship cues between each pair of connected vertices is increased from one to $K$) for various graph analysis tasks. These results also provide an evidence that it is not enough to only use a single value to describe the relationship (the edge) between each pair of connected vertices whose representations are multi-dimensional.

\textbf{Benefits of the TTP-MEFG:} Finally, it is clear that the system consisting of both TTP and MEFG modules frequently leads to further improvements over the systems that individually use either of them, i.e., the system consisting of both TTP and MEFG achieved the best result in five out of eight graph and vertex classification cases on graph datasets (Table \ref{tb:Ablation_graph}). Meanwhile, Table \ref{tb:Ablation_nongraph_Res} and Table \ref{tb:Ablation_nongraph_Swin} show that using the entire proposed pipeline also leads to improvements for either only using TTP or MEFG on non-graph datasets, regardless of the employed backbone. 
These results suggest that both task-specific topology and task-specific multi-dimensional edge features produced by TTP and MEFG modules can enhance the generated graph representation from the original graphs or latent representations. More importantly, they enhance graphs in different ways by encoding task-specific and complementary cues in the generated graph, i.e., the benefits brought by TTP and MEFG are complementary to each other.

\textbf{Edge representations for link prediction tasks:} The standard way to predict a graph link (i.e., presence of edges) between a pair of vertices is to feed their features to the classifier. The proposed MEFG module allows to further learn a multi-dimensional edge feature to describe the task-specific relationship between each pair of vertices, regardless of whether they are connected or not. As shown in Table \ref{tb:link prediction}, we feed three types of inputs to the MLP classifier for predicting links:(i) the vertex features which are updated with the help of our multi-dimensional edge feature, i.e., they are produced from the last GAT/GatedGCN layer only; (ii) the edge feature (deep-learned multi-dimensional edge feature produced from the last GAT/GatedGCN layer) only; and (iii) both edge and vertex features produced from the last GAT/GatedGCN layer of our system. It can be observed that the link prediction results achieved by vertex features or edge feature produced by the proposed approach show clear advantages over the original pre-defined vertex features on both datasets. This indicates that the proposed approach can learn strong edge representations that can not only directly describe the task-specific relationship between vertices, but also further help to learn more task-specific vertex representations. Although combining vertex features and multi-dimensional edge features show the best performances on both graph datasets, combining them does not always provide the best results on non-graph dataset. This is due to the fact that the edge features of non-graph graph representations are fully learned from their vertices, and thus they may contain redundant information encoded in the corresponding vertices.

\begin{table}[t]
\caption{Link prediction results achieved for different settings on both graph and non-graph datasets, where baselines denote the original vertex features provided by graph datasets (TSP and COLLAB) or pre-trained Swin-Transformer (BP4D and DISFA).}
\label{tb:link_prediction}
\centering
\resizebox{\linewidth}{!}{
\begin{tabular}{lcccc}
\toprule
\multirow{2}*{\textbf{Method}} &\multicolumn{2}{c}{\textbf{Graph datasets}} &\multicolumn{2}{c}{\textbf{Non-graph datasets}} \\
\cline{2-5} 
{} & \textbf{TSP} & \textbf{COLLAB} &\textbf{BP4D} & \textbf{DISFA} \\
\toprule
\multicolumn{5}{l}{\textsl{(A) Predictor: GatedGCN} } \\ 
Baseline  &0.808 &52.64 & 58.2 & 65.1\\
Vertices  &0.826 &54.17 & 60.6 & 68.3  \\
Edge  &0.813 & 53.12 & \textbf{61.1} & \textbf{71.7}  \\
Vertices+Edge &\textbf{0.830} &\textbf{54.76} & 60.8 & 71.5   \\
\toprule
\multicolumn{5}{l}{\textsl{(B) Predictor: GAT} } \\ 
Baseline  &0.671 &51.50  & 58.2  & 67.2 \\
Vertices  &0.792 &53.09 & \textbf{61.8} & 71.8 \\
Edge  &0.725 &52.34 & 60.0  & \textbf{72.6}  \\
Vertices+Edge  &\textbf{0.807} &\textbf{53.68} & 61.7 & 71.4   \\
\toprule
\end{tabular}}
\label{tb:link prediction}
\end{table}

\begin{figure}
	\centering
	\subfigure[Example multi-dimensional edge features on BP4D dataset. ]{
		\includegraphics[width=9cm]{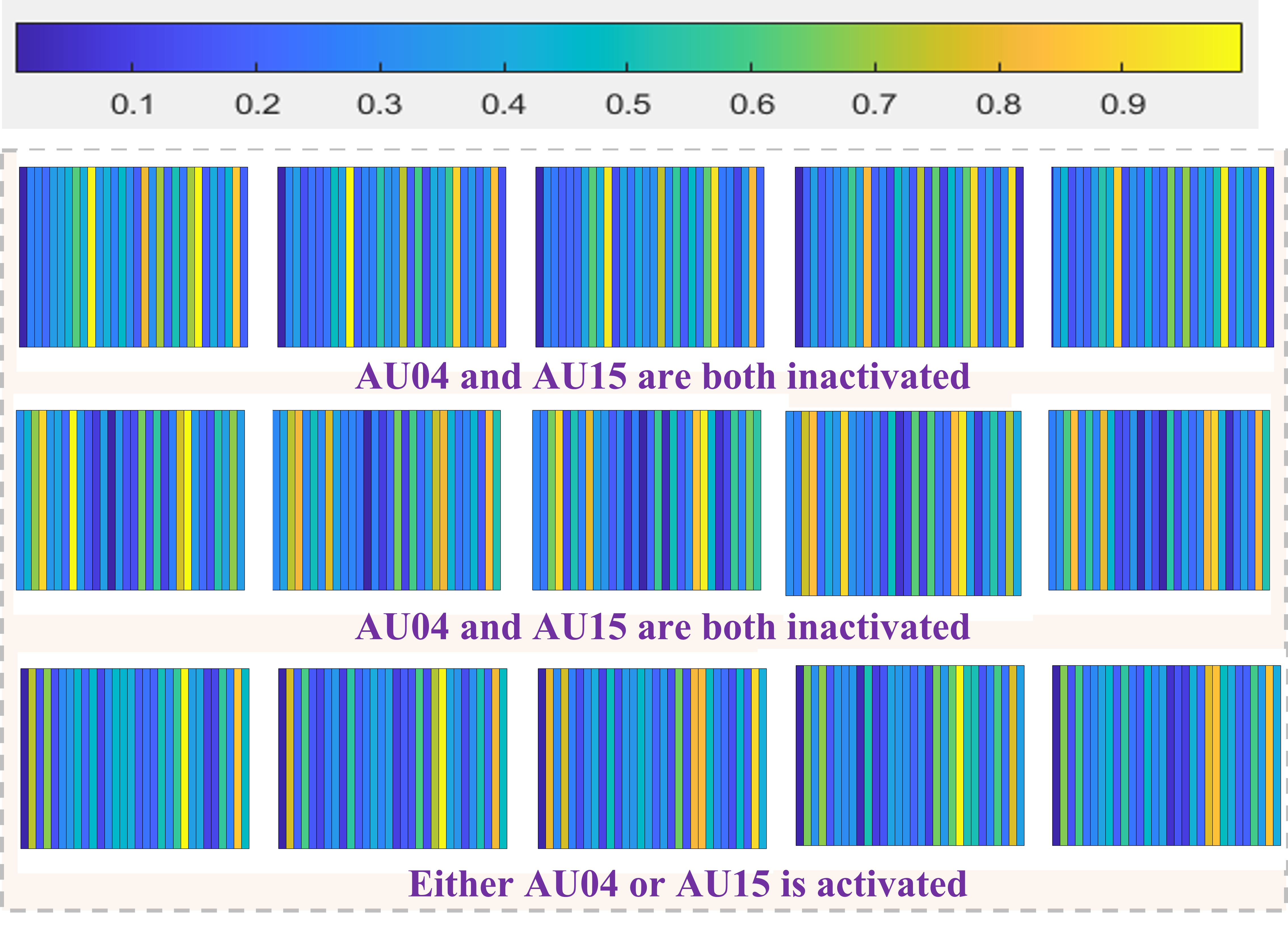}}
	\subfigure[Example multi-dimensional edge features on TSP dataset.]{
		\includegraphics[width=9cm]{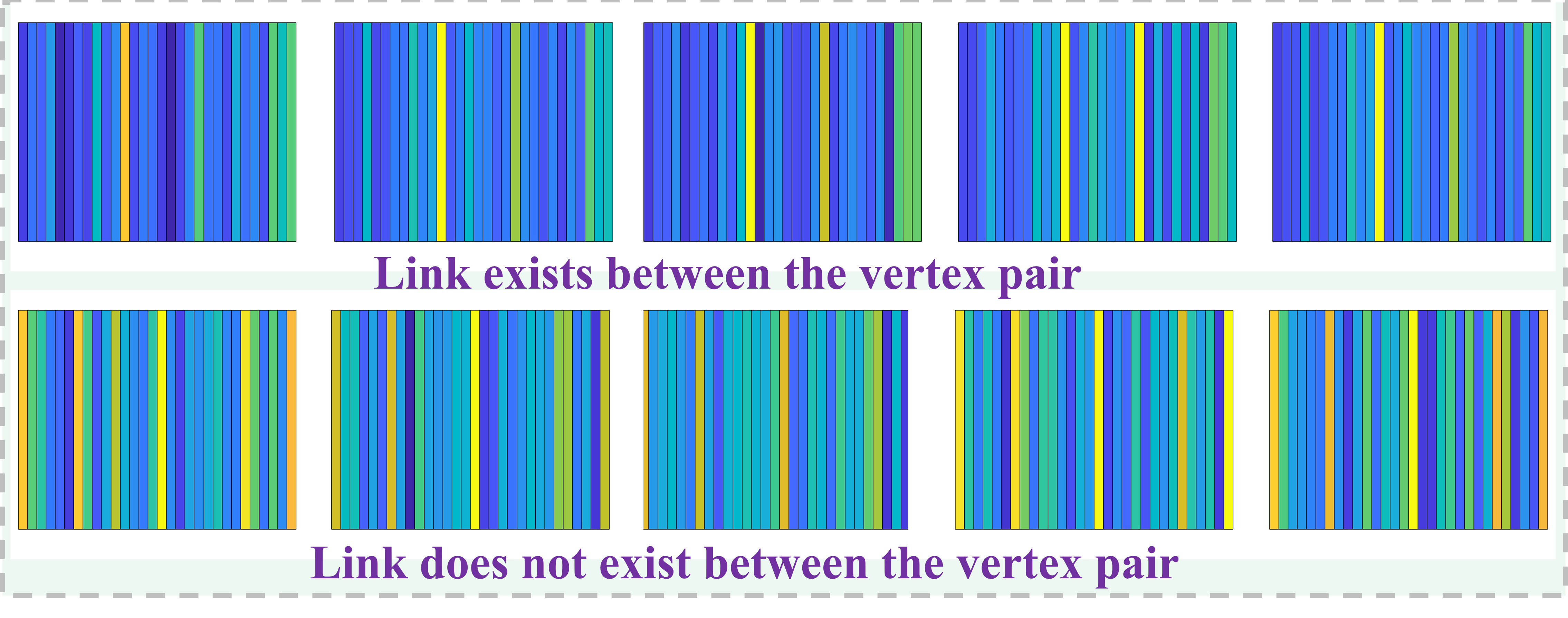}}
	\caption{Visualization of normalized example multi-dimensional edge features produced by our MEFG on graph and non-graph datasets.} 
    \label{fig:edge}
\end{figure}

\begin{figure}[htb]
  \centering
  \includegraphics[width=1\columnwidth]{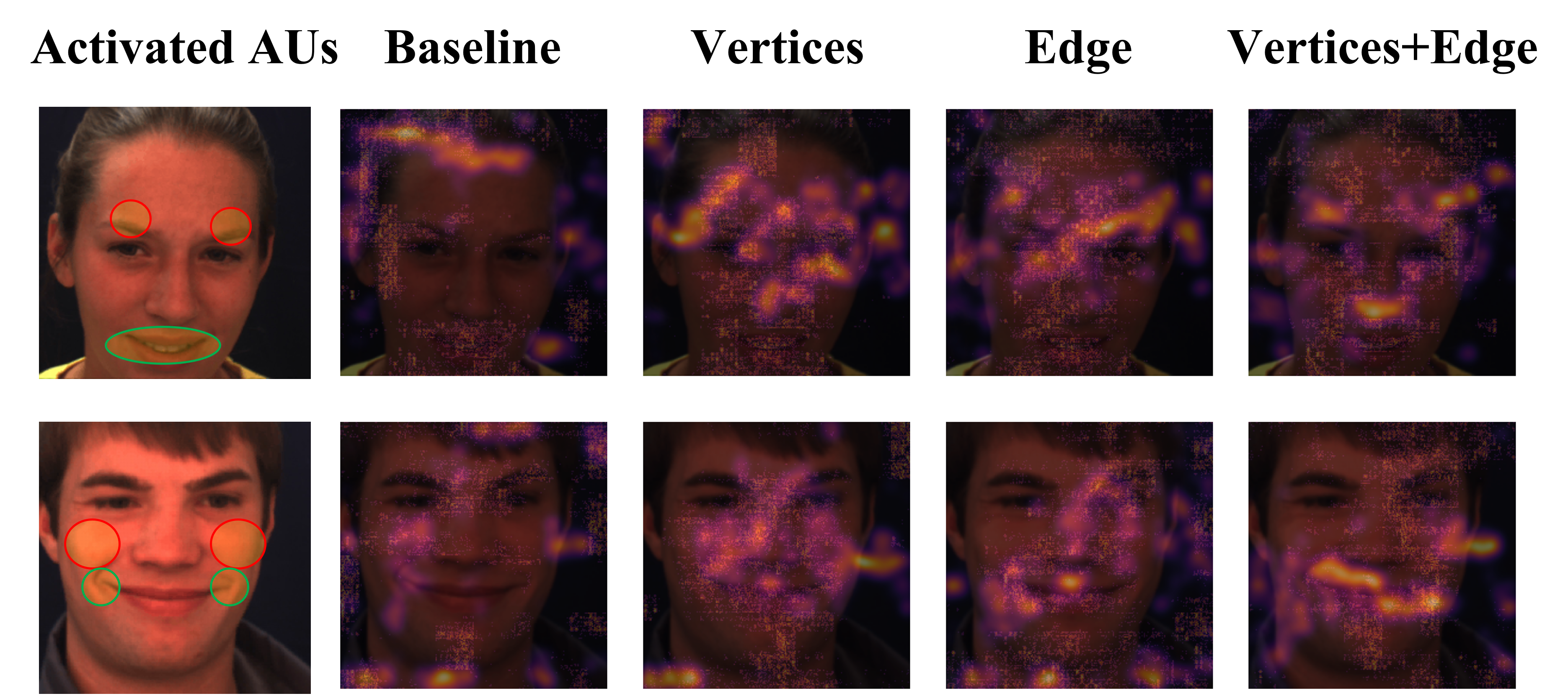}
  \caption{Saliency maps of the AU co-occurrence detection task generated from the baseline vertex features, our vertex features, our edge features as well as our vertex+edge features on DISFA dataset. The upper face has two activated AUs: AU4 (eyebrow lowerer) and AU25 (lips part) while the lower face also has two activated AUs: AU6 (check raiser) and AU12 (lip corner puller).}
\label{fig:Visualization_co_occurrence}
\end{figure}

\textbf{Discussion:} We found that the improvements brought by MEFG are usually larger than the TTP on graph datasets (i.e., with average 1.53\% and 1.03\% improvements for GatedGCN and GAT-based systems, respectively). We assume that this is because most pre-defined graphs provided in the graph datasets already have reliable typologies, and thus TTP can only provide limited improvements. Meanwhile, MEFG can largely enhance pre-defined graphs by assigning a task-specific multi-dimensional edge representation for each pre-defined edge, as these multi-dimensional edges contain much more task-specific information than the pre-defined single-value edges. In contrast, TTP is more crucial for building graph representations from non-graph data. The systems that only used TTP achieved average 0.88\%, 0.65\%, 0.36\%, and 0.27\% advantages over the systems that only used MEFG (i.e., ResNet-MEFG-GatedGCN, ResNet-MEFG-GAT, Swin-MEFG-GatedGCN and Swin-MEFG-GAT). This is because the TTP plays a key role in defining the graph representations for non-graph samples. Since there is no pre-defined topology and vertex features for each non-graph sample, it takes the full responsibility for generating task-specific typologies and vertex features for them, i.e., the learned graph representations are more powerful than the latent representation produed from the backbone for various tasks. However, directly using MEFG without the TTP would result in the vertex features directly coming from the GD module, and the topology of the graph are manually defined (fully connected in this paper), both of which are not optimal. We assume that only assigning a deep-learned multi-dimensional representation to each edge can not compensate the task-specific cues ignored in such topology and vertex features. More importantly, the two modules are complementary to each other, making each enhanced graph to have not only task-specific topology but also task-specific edge representations, and thus the combination of them results in the best performances on various datasets. We also visualized some example multi-dimensional edge features learned by our approach in Fig. \ref{fig:edge}, where edge features are clear different when they represent different relationships. Meanwhile, Fig. \ref{fig:Visualization_co_occurrence} show that features extracted by all three settings of our approach are related to the facial regions of activated AUs as well as other potentially related facial regions. In contrast, features extracted by the baseline either ignore the related facial regions or focus on less-related facial regions. Moreover, using both vertex and edge features learned by our approach allows the features to be strictly extracted from the facial regions of activated AUs.



\begin{figure*}[htb]
  \centering
  \includegraphics[width=2\columnwidth]{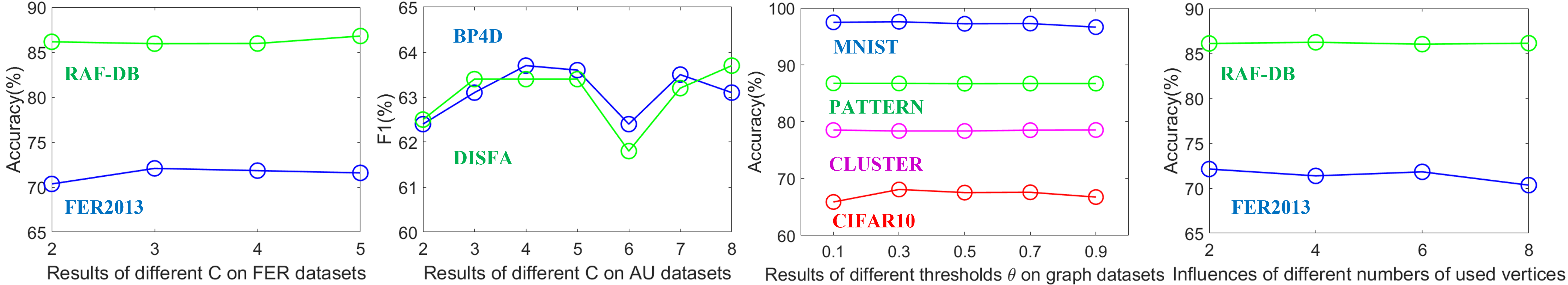}
  \caption{Parameter sensitivity analysis results, where the performance of our GRATIS is robust to main parameter settings.}
\label{fig:sensitivity_analysis}
\end{figure*}

\subsubsection{Parameter sensitivity analysis}
\label{subsec:Parameter}

\noindent We specifically analyze three main parameters of the proposed approach on multiple datasets in Fig. \ref{fig:sensitivity_analysis}, which are: (i) the '$C$' employed in Eqa. \ref{eq:C-nearest}, which decides the number of neighbours that are connected by each vertex for non-graph data-based experiments; (ii) the threshold $\theta$ of the Eqa \ref{eq:ttp_graph} in the TTP module, which decides the number of vertices that are connected by each vertex for graph-based experiments; and (iii) the number of vertices used for describing each non-graph sample for graph classification tasks (i.e., the number of vertices can not be changed for vertex and link prediction tasks). It is clear that the results achieved by our approach under different parameter settings are stable. The performance variations on four non-graph datasets are less than 3\% when various '$C$' values are employed. In particular, the average AU recognition F1 results of different $C$ values achieved on both BP4D and DISFA datasets consistently outperform most existing methods. Similarly, different $\theta$ values as well as different vertex numbers defined for each graph representation, also only have very small influences on the results of different graph analysis tasks. As a result, we can conclude that \textbf{our approach is robust to main parameters}, i.e., it is easy to apply and extend our approach for various other proper tasks.

\section{Conclusion and discussion}

\noindent In this paper, we propose the first generic and plug-and-play framework that can generate an enhanced graph representation with task-specific topology and multi-dimensional edge features, from any arbitrary graph or non-graph data. The proposed framework consists of three main modules: (i) \textbf{Graph Definition module} that builds a raw graph representation from the input data; (ii) \textbf{Task-specific Topology Prediction module} that assigns a task-specific adjacency matrix to define edges' presence of the graph representation, as well as learn a set of task-specific vertex features for non-graph data; and (iii) \textbf{Multi-dimensional Edge Feature Generation module} that produces a task-specific multi-dimensional edge feature to describe each presented edge. 

We conducted a set of experiments to systematically evaluate the proposed framework on various graph and non-graph datasets that can be treated as different graph analysis tasks (i.e., graph classification, vertex classification and link prediction). Consequently, several conclusions can be drawn from the results obtained: (i) the proposed framework can automatically build a strong graph representation for any pre-defined graph or non-graph data, which frequently generates superior performances for various tasks; (ii) both TTP and MEFG can encode task-specific cues for the graph representation, which are represented by task-specific topology/vertex features and task-specific multi-dimensional edge features. The task-specific cues represented by them are complementary, as the combination of them produced the best results; (iii) multi-dimensional edge feature is essential for describing graph edges when vertex features have multiple dimensions, as it achieved better performance than different single-dimensional edge features. Our MEFG is the first approach that can effectively deep learn task-specific multi-dimensional edge features for any arbitrary data (e.g., single-value edge graph, non-graph data, etc.); (iv) the proposed framework is robust to the change of main parameters (Sec. \ref{subsec:Parameter}); and
(v) Our framework is robust to various backbones and GNN predictors, as it consistently improved the performance under different backbone and GNN predictor settings.


\textbf{Limitations:} Although GRATIS is the first general framework that deep learns both task-specific topology and multi-dimensional edge features for any arbitrary data, the main limitation is that the framework still contains several modules and relatively large number of CNN/GNN/attention layers. Consequently, it has a certain number of weights to be optimized. Moreover, since the final adjacency matrix produced by our TTP module is not directly involved in the propagation and back-propagation process, only the probability adjacency matrix $\mathcal{\hat{A}}^{\text{prob}}$ or a vertex-decided adjacency matrix $\mathcal{\hat{A}}^{\text{V}}$ are task-specific. 

\textbf{Future work:} Due to the limited spaces and resources, our experiments were not able to evaluate the proposed approach on all available graph and non-graph datasets (e.g., the other six datasets proposed in \cite{dwivedi2020benchmarking} or other famous graph datasets). Therefore, our future work will focus on: (i) making the framework more compact and lightweight; (ii) allowing the final adjacency matrix to be optimized end-to-end; and (iii) reporting the results on more datasets in the future version.


%



\ifCLASSOPTIONcompsoc
  \section*{Acknowledgments}
  
\noindent\textbf{Funding:} The work of Siyang Song and Hatice Gunes is funded by the EPSRC/UKRI under grant ref. EP/R030782/1.

\noindent\textbf{Open Access Statement:} For the purpose of \textit{open access}, the authors have applied a Creative Commons Attribution (CC BY) license to any Accepted Manuscript version arising. 

\noindent\textbf{Data Access Statement:} This study involves secondary analyses of existing datasets, that are described and cited in the text. Licensing restrictions prevent sharing of the datasets. 
  
\else

  \section*{Acknowledgment}
\fi

\ifCLASSOPTIONcaptionsoff
  \newpage
\fi

\bibliographystyle{IEEEtran}

\bibliography{ref}





\end{document}